%% file: main.tex
\documentclass[10pt,twocolumn,letterpaper]{article}

\usepackage[pagenumbers]{cvpr} % To force page numbers, e.g. for an arXiv version
\usepackage{graphicx}
\usepackage{subcaption}
\usepackage{enumitem} 
\usepackage[table]{xcolor}
\usepackage{pifont}
\usepackage[most]{tcolorbox}

\usepackage{booktabs}    % \toprule, \midrule, \bottomrule
\usepackage{multirow}
\usepackage{makecell}    % \shortstack
\usepackage{arydshln}    % \hdashline
\usepackage{graphicx}
\usepackage{caption}
\usepackage{subcaption}
\definecolor{softlavender}{RGB}{245, 240, 255}
\definecolor{softblue}{RGB}{230,242,255}
\definecolor{tablegray}{gray}{0.95}

\definecolor{pastelYellow}{HTML}{FFFADC}
\definecolor{pastelMint}{HTML}{DCEFE6}
\definecolor{pastelLavender}{HTML}{EDE8FF}
\definecolor{pastelPeach}{HTML}{FFEBD9}

\newcommand{\xmark}{\text{\ding{55}}}  % pifont
\definecolor{darkgreen}{RGB}{0,127,0}
\definecolor{darkred}{RGB}{200,0,0}
\def\greencheckmark{\textcolor{darkgreen}{\checkmark}}
\def\redxmark{\textcolor{darkred}{\xmark}}

\input{preamble}

\definecolor{cvprblue}{rgb}{0.21,0.49,0.74}
\usepackage[pagebackref,breaklinks,colorlinks,allcolors=cvprblue]{hyperref}
\usepackage{caption}
\usepackage{float}
\usepackage{booktabs}

\def\paperID{5090} % *** Enter the Paper ID here
\def\confName{CVPR}
\def\confYear{2026}

\title{MindPower: Enabling Theory-of-Mind Reasoning in VLM-based Embodied Agents}

\author{
    Ruoxuan Zhang$^1$ \quad Qiyun Zheng$^1$ \quad Zhiyu Zhou$^1$ \quad Ziqi Liao$^1$ \quad Siyu Wu$^1$ \\
    Jian-Yu Jiang-Lin$^2$ \quad Bin Wen$^1$ \quad Hongxia Xie$^{1, *}$ \quad Jianlong Fu$^3$ \quad Wen-Huang Cheng$^2$ \\[6pt]
    $^1$Jilin University \qquad
    $^2$National Taiwan University \qquad
    $^3$Microsoft Research Asia\\[6pt]
    \small{$^*$Corresponding author: \texttt{hongxiaxie@jlu.edu.cn}}
}
\begin{document}
\twocolumn[{%
\renewcommand\twocolumn[1][]{#1}%
\maketitle
\begin{center}
    \centering
    \includegraphics[width=1\textwidth,height=10.3cm]{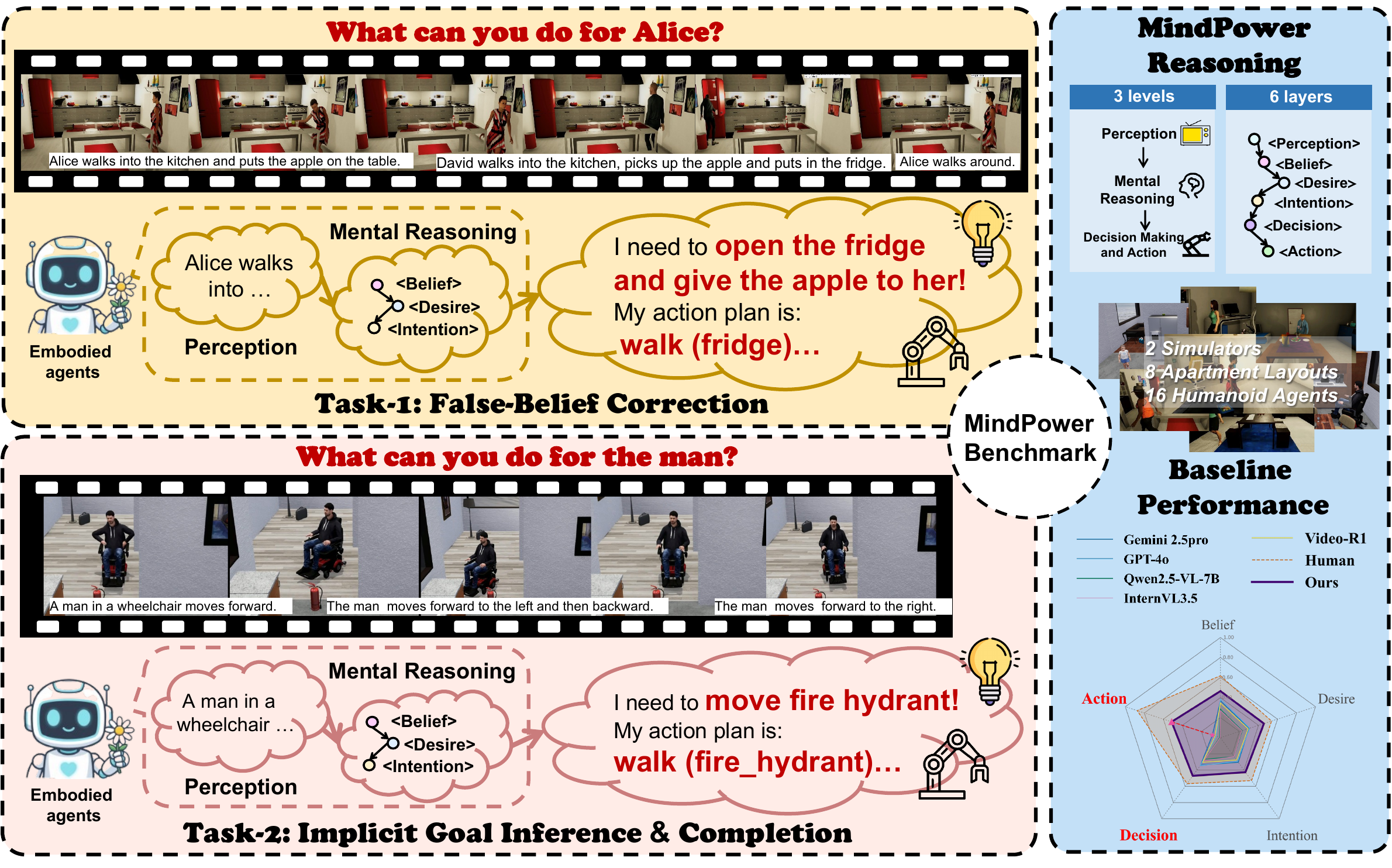}
    \captionof{figure}{ \textbf{MindPower Benchmark Overview.} We evaluate Robot-Centric ToM through two tasks: \textbf{False-Belief Correction} and \textbf{Implicit Goal Inference \& Completion}, assessing whether VLM-based embodied agents can generate correct decisions and actions.
We further propose the \textbf{MindPower Reasoning Hierarchy}, comprising three levels and six layers.
Existing VLMs perform poorly across layers, especially in action reasoning, while our model shows substantial improvements.
A detailed example is provided in Supp. Sec.~B.}
    \label{fig:overview}
\end{center}%
}]
\begin{abstract}
Theory of Mind (ToM) refers to the ability to infer others’ mental states, such as beliefs, desires, and intentions.  
Current vision–language embodied agents lack ToM-based decision-making, and existing benchmarks focus solely on human mental states while ignoring the agent’s own perspective, hindering coherent decision and action generation.
To address this, we propose MindPower, a Robot-Centric framework integrating Perception, Mental Reasoning, Decision Making and Action. 
Given multimodal inputs, MindPower first perceives the environment and human states, then performs ToM Reasoning to model both self and others, and finally generates decisions and actions guided by inferred mental states. 
Furthermore, we introduce Mind-Reward, a novel optimization objective that encourages VLMs to produce consistent ToM Reasoning and behavior. Our model outperforms GPT-4o by 12.77\% in decision making and 12.49\% in action generation. Benchmark will be available at \href{https://zhangdaxia22.github.io/MindPower/}{\textcolor{blue}{https://zhangdaxia22.github.io/MindPower/}}.
\end{abstract}    

\section{Introduction}
\label{sec:intro}

Understanding human mental states is a prerequisite for genuine human–agent collaboration. Unlike conventional embodied systems that merely execute explicit commands, next-generation agents must reason about what humans believe, desire, and intend, and act proactively on that understanding~\cite{fung2025embodied}. This requires an explicit mental reasoning mechanism based on the human Theory of Mind (ToM)~\cite{ leslie2004core,onishi200515,frith2005theory}, which can be formalized by the Belief–Desire–Intention (BDI) framework~\cite{rao1995bdi}. In BDI, humans perceive the world and others’ behaviors, form beliefs about the environment, derive desires that encode goals, and generate intentions that guide actions, reflecting ToM Reasoning. This raises the question: \textit{Can embodied agents reason and act in a similar ToM-consistent manner?}

We formalize this cognitive process into three progressive levels of embodied intelligence.  
(1)~\textbf{Perception}: understanding human behaviors and environmental contexts via vision–language reasoning.  
(2)~\textbf{Mental Reasoning}: inferring human beliefs, desires, and intentions, as demonstrated in first- and second-order ToM Reasoning tasks.  
(3)~\textbf{Decision Making and Action}: reasoning about one’s own beliefs and intentions to make autonomous, goal-directed decisions and provide proactive assistance.  
This three-level hierarchy bridges perception and intention, paving the way toward truly collaborative human–AI interaction.

Despite rapid progress in Vision–Language Models (VLMs), a fundamental gap remains in embodied intelligence. As shown in the bottom right of Fig.~\ref{fig:overview}, current VLMs such as Gemini~\cite{comanici2025gemini}, GPT~\cite{achiam2023gpt}, and Qwen-VL~\cite{bai2025qwen2} excel at perception but remain largely reactive. They can describe what they see, yet fail to reason about what humans believe, desire, or intend.
Existing Theory-of-Mind (ToM) benchmarks~\cite{shi2025muma,jin2024mmtom,mao2024bdiqa} have endowed VLMs with certain mental reasoning abilities, but they are limited to reasoning about the mental states of humans appearing in the video. They do not build ToM Reasoning from their own perspective, which prevents VLMs from learning to make decisions and generate actions.

We address this gap through a \textbf{Robot-Centric Perspective}, which enables VLMs to reason simultaneously about their own mental states and those of humans, forming a continuous and interpretable ToM Reasoning loop. Inspired by frameworks such as LLaVA-CoT~\cite{xu2025llava} and Visual-RFT~\cite{liu2025visual}, we further design the Robot-Centric \textbf{MindPower Reasoning Hierarchy}, which connects ToM Reasoning with decision making and action generation. It structures reasoning into three levels and six layers: from \texttt{<Perception>} (Perception), through \texttt{<Belief>}, \texttt{<Desire>}, and \texttt{<Intention>} (Mental Reasoning), to \texttt{<Decision>} and \texttt{<Action>} (Decision Making and Action).

To realize this goal, we introduce the \textbf{MindPower Benchmark}. An overview of our benchmark, reasoning hierarchy, and experiments is shown in Fig.~\ref{fig:overview}.
MindPower comprises two core embodied reasoning tasks:
(1)~\textbf{False-Belief Correction}, which examines whether an embodied agent can detect and resolve a human’s mistaken belief; and
(2)~\textbf{Implicit Goal Inference \& Completion}, which tests whether the agent can infer a hidden goal and assist in achieving it.
We construct 590 scenarios across two interactive home-environment simulators, each containing multimodal observations and object-manipulation activities that reflect everyday embodied reasoning challenges.

Further, to enhance reasoning consistency across these layers, we propose Mind-Reward, a reinforcement-based optimization framework that aligns intermediate ToM states with final actions, promoting Robot-Centric, continuous reasoning.

Our contributions are threefold:
\begin{itemize}
    \item Robot-Centric Perception Benchmark for mental-state-grounded action. 
  MindPower links mental reasoning with embodied action through two tasks: \textit{False-Belief Correction} and \textit{Implicit Goal Inference \& Completion}, across 590 interactive home scenarios, evaluating agents’ ability to infer, make decisions, and assist.
    \item Unified MindPower Reasoning Hierarchy bridging perception and action.
    The MindPower Reasoning Hierarchy structures reasoning across three levels and six layers, providing a standardized way to evaluate how perception leads to action.    
    \item Reinforcement optimization for consistent ToM Reasoning.
   Mind-Reward aligns intermediate reasoning states with final actions, promoting coherent Robot-Centric reasoning.  
    With this optimization, our model surpasses GPT-4o by 12.77\% in decision accuracy and 12.49\% in action generation.
\end{itemize}

\section{Related Work}
\label{sec:relatedwork}
\noindent \textbf{Theory of Mind Benchmark.} 
Early ToM benchmarks relied on narrative text to infer beliefs, desires, and intentions~\cite{he2023hi,kim2023fantom,gandhi2023understanding,le2019revisiting,wilf2024think,cheng2024egothink,Gao_2025_ACMMM, yangembodiedbench,jung2024perceptions}, but lacked multimodal grounding.
Subsequent multimodal benchmarks introduced videos or images depicting story-based social scenarios to support richer mental-state inference~\cite{jin2024mmtom,shi2025muma,fan2025somi,li2025black,du2024constrained,mao2024bdiqa,villa2025moments,zhang2025autotom}.
However, most adopt multiple-choice or short-answer formats and focus on role-level or factual queries, offering limited support for open-ended, real-world reasoning where agents must update beliefs and act continuously.
Although datasets such as MuMA-ToM~\cite{shi2025muma} and MMToM-QA~\cite{jin2024mmtom} explore false-belief understanding or implicit goal inference, they still do not support dynamic reasoning processes that involve belief correction, assistance-oriented behavior, or proactive decision-making, which are essential for autonomous embodied agents.

\noindent \textbf{VLMs-based Embodied Agents.} 
Embodied agents have been developed to perform tasks autonomously by decomposing complex goals into multiple subtasks and executing them step by step~\cite{chen2023egoplan,sermanet2024robovqa,luo2025robobench,li2024mmro,zhang2025vlabench}. For example, PaLM-E~\cite{palme} demonstrates that large embodied models can perform high-level task planning by integrating visual and linguistic cues.
Some benchmarks further support multi-agent collaboration, enabling agents to observe each other or even human partners to coordinate goals and actions~\cite{zhang2024combo, puig2020watch,jenamani2025feast,wang2025strangers,ding2024atom}. For example, RoboBench~\cite{luo2025robobench} allows agents to decompose high-level goals into subgoals for sequential execution, while Smart-Help~\cite{cao2024smart} focuses on achieving comfortable human–robot interaction by balancing human comfort and task efficiency.
However, these systems still depend on predefined goals or imitation signals and lack self-perspective mental reasoning. As highlighted in Mindblindness~\cite{baron1997mindblindness}, social intelligence requires inferring others’ mental states and acting upon those inferences, a capability missing from current embodied benchmarks. They do not evaluate first- or second-order belief reasoning, which is crucial for autonomous and socially grounded decision-making.
Even robotic setups that incorporate hidden-belief modeling, such as AToM-Bot~\cite{ding2024atom}, cover only narrow goal spaces and provide limited task diversity, falling short of comprehensive ToM evaluation.

\section{MindPower Benchmark}

\subsection{Problem Definition and Cognitive Inspiration}

% The Theory of Mind (ToM) framework~\cite{rao1995bdi} models human decision-making as a three-level hierarchy. A person first forms desires based on their beliefs about the environment, and then selects and commits to specific intentions to act on those desires. 
The Theory of Mind (ToM) framework~\cite{rao1995bdi} models human decision-making through a Belief–Desire–Intention hierarchy: individuals form desires from their beliefs and commit to intentions that drive actions. 
Building on this cognitive structure, we introduce the \textbf{MindPower Benchmark}, which includes a unified reasoning hierarchy (i.e., MindPower Reasoning Hierarchy), the curated MindPower dataset, and comprehensive evaluation metrics.

\begin{figure}[t]
  \centering
  \includegraphics[width=1\linewidth]{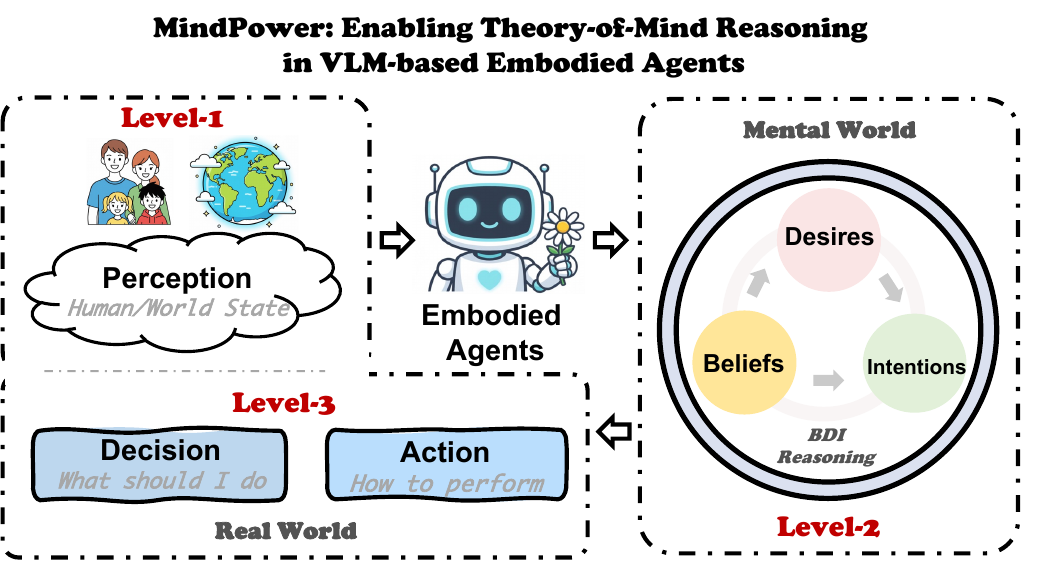} 
  \caption{\textbf{MindPower Reasoning Hierarchy.} The agent first receives multimodal input, then performs mental reasoning to form beliefs, desires, and intentions, and finally makes decisions and generate action plan based on this reasoning.}
  \label{fig:mindcot}
\end{figure}

Specifically, as shown in Fig.~\ref{fig:mindcot}, the \textbf{MindPower Reasoning Hierarchy} extends the embodied decision-making process into six layers organized across three levels, each reflecting how an embodied agent perceives, reasons, and acts within its environment.

\noindent \textbf{Level-1: Perception.}  
\begin{itemize}
 \item \texttt{<Perception>} — The agent observes the environment through vision or other sensory inputs. This step answers \textit{“What is happening now?”}
\end{itemize}

\noindent \textbf{Level-2: Mental Reasoning.}  
\begin{itemize}
    \item \texttt{<Belief>} — Reasoning about both human and environmental states based on perception. Unlike first-order belief, which reflects only the agent’s own understanding, our framework models \textbf{second-order belief}: the agent infers not only its own beliefs but also what it predicts humans in the scene believe.

    \item \texttt{<Desire>} — A preferred state or goal derived from the agent’s beliefs. For an embodied helper agent, desires are shaped by the goal of assisting humans and determining what assistance is needed and why.  
    \item \texttt{<Intention>} — A concrete commitment to act, formed based on the agent’s beliefs and desires.
\end{itemize}

\noindent \textbf{Level-3: Decision Making and Action.}  
\begin{itemize}
\item \texttt{<Decision>} — The choice or plan the embodied agents makes to fulfill the intention.
\item \texttt{<Action>} — The action execution sequence, where the embodied agent enacts its decisions through high-level atomic operations in the form of \texttt{action (object)}, such as \texttt{open (fridge)} or \texttt{pick\_up (milk)}.
\end{itemize}

\begin{figure*}[t]
  \centering
  \includegraphics[width=1\linewidth]{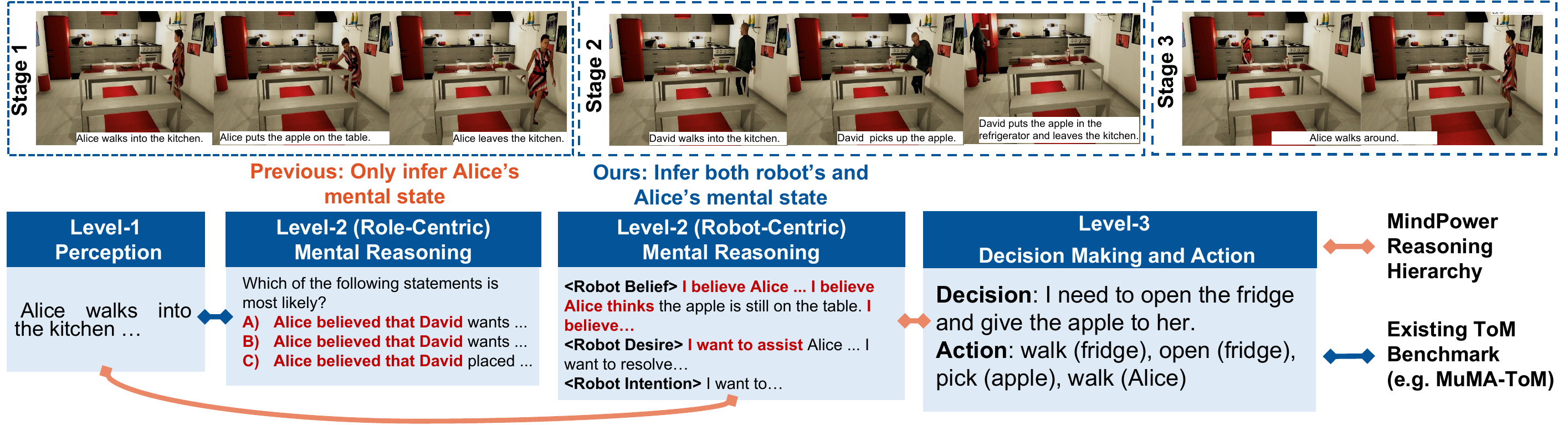} 
  \caption{\textbf{Robot-Centric MindPower Reasoning Hierarchy.}
% We use an example to compare our Robot-Centric MindPower Reasoning Hierarchy with existing ToM Benchmarks.  
Existing benchmarks, such as MuMA-ToM, include only Stage~1 and Stage~2 of the video, and focus solely on inferring the mental reasoning of the human (Alice) in the input video. Our dataset additionally includes Stage~3, where Alice returns to search for the item. Moreover, in Level-2 (Mental Reasoning) of MindPower, we infer the mental reasoning of both the embodied agent and the human, whereas existing ToM Benchmarks only infer the role’s mental state through multiple-choice questions.
Detailed example is provided in Sec.~B of the Supplementary Material. }
  \label{fig:compare}
\end{figure*}

\begin{table*}[t]
\scriptsize
\centering
\caption{
\textbf{Comparison of Theory-of-Mind (ToM) Benchmarks.} ``MCQ" denotes Multiple Choice Question. 
 “Level-3 Ability” indicates whether each dataset involves 
\textit{False-Belief Correction}, \textit{Implicit Goal Inference \& Completion}, 
and \textit{Decision Making and Action level}.
}
\label{tab:tom-datasets}
\setlength{\tabcolsep}{4pt}
\begin{tabularx}{\linewidth}{lccccccc}
\toprule
\textbf{Dataset} & \textbf{Modality} & \textbf{Output} &
\textbf{Agent Type} & \textbf{Perspective} &
\textbf{Format} & \textbf{Scale} &
\textbf{Level-3 Ability} \\
\midrule
Hi-ToM~\cite{he2023hi} & Text & Belief &
- & Role-Centric & MCQ & 1,800 stories & \redxmark  \\
BigToM~\cite{gandhi2023understanding} & Text & Belief, Role's action &
- & Role-Centric & MCQ & 5,000 text items & \redxmark  \\

FANToM~\cite{kim2023fantom} & Text & Belief &
- & Role-Centric & MCQ & 256 stories & \redxmark  \\
MuMA-ToM~\cite{shi2025muma} & Video, Text & Belief, Goal &
Virtual Human & Role-Centric & MCQ & 225 examples & \redxmark  \\
MMToM-QA~\cite{jin2024mmtom} & Video, Text & Belief, Goal &
Virtual Human & Role-Centric & MCQ & 134 videos & \redxmark  \\
% BDIQA~\cite{mao2024bdiqa} & Video, Text &  &
% Virtual Human & Role-centric & MCQ &  & \xmark  \\
GridToM~\cite{li2025black} & Video, Text & Belief &
Grid-world Agent & Role-Centric & MCQ & 1,296 videos & \redxmark  \\
SoMi-ToM~\cite{fan2025somi} & Video, Image & State, Goal, Behavior &
Minecraft Roles & Role-Centric & MCQ & 35 videos / 363 images & \redxmark  \\
\hdashline
\textbf{Ours} & \textbf{Video, Text} &
\makecell{\textbf{Perception, Belief, Desire,}\\ \textbf{Intention, Decision, Action}} &
\textbf{Virtual Human} & \textbf{Robot-Centric} &
\textbf{Open-Ended} & \textbf{590 examples} &
\greencheckmark \\
\bottomrule
\end{tabularx}
\end{table*}

\subsection{Mindpower Dataset Collection}
Based on proposed MindPower Reasoning Hierarchy, we propose MindPower Dataset.

\noindent\textbf{Dataset Collection Principles.}
We construct the dataset based on three principles: 
(1) \textbf{Realism:} scenarios and events should be plausible in the real world. 
(2) \textbf{BDI Consistency:} 
% each sample must preserve a coherent hierarchical structure from \texttt{<Perception>} to \texttt{<Action>}. Intermediate layers should remain consistent and non-contradictory.  
each sample preserves a coherent \texttt{<Perception>} to \texttt{<Action>} hierarchy with logically consistent intermediate states.
(3) \textbf{Diversity under simulator constraints:} 
within simulator constraints, we include varied scenes, roles, and goals to ensure diversity while maintaining feasible simulation and annotation.
% within the limits of current simulators, we include varied scenes, agent roles, and goal types to increase diversity while maintaining feasibility for simulation and annotation.

% \noindent \textbf{Task Design.}
% To build an effective MindPower dataset, we design two primary task types: \textbf{False-Belief Correction} and \textbf{Implicit Goal Inference}.  
% In the False-Belief Correction task, when a human character holds an incorrect belief about the location of an object, the embodied agent is expected to recognize the mismatch, retrieve the object from its correct location, and hand it to the human.  
% In the Implicit Goal Inference, the agent must infer the user’s latent goal even when it is not explicitly stated. 
% In addition, we design scenarios involving individuals with specific needs, such as a wheelchair user and a 1.2-meter-tall child. These cases allow the agent to demonstrate adaptive reasoning and assistance behaviors for users with mobility or reach limitations. 

 \noindent \textbf{Task Design.} 
 % To construct an effective MindPower dataset, we design two primary task types:
% \textbf{False-Belief Correction} and \textbf{Implicit Goal Inference and Completion}.
% The first type focuses on the embodied agent's ability to detect and correct a human’s false belief, 
% while the second evaluates the embodied agent’s capacity to infer unstated user intentions.
We define two core task types for constructing the MindPower dataset: \textbf{False-Belief Correction} and \textbf{Implicit Goal Inference \& Completion}.
The former evaluates whether an embodied agent can detect and correct a human’s mistaken belief about the environment (e.g., misjudged object locations).
The latter tests the agent’s ability to infer unstated intentions from subtle behavioral cues, such as searching or repeated failed attempts.
For example, when the human starts rummaging through drawers or walking around to search after completing several actions, the agent should reason that the human is looking for a specific target object.
% In the False-Belief Correction task, a human holds an incorrect belief about the location of an object. 
% The embodied agent must recognize this mismatch and help correct the false belief (e.g., by locating the object and handing it to the human). 
% In the Implicit Goal Inference and Completion task, the embodied agent needs to infer the human’s implicit goal. 
We further incorporate special-needs scenarios (e.g., wheelchair users or children), enabling evaluation of assistive behaviors under mobility and reach constraints.

Different from ToM benchmarks such as MuMA-ToM~\cite{shi2025muma} and MMToM-QA~\cite{jin2024mmtom}, our task explicitly models the moment when belief contradictions arise. As shown in Fig.~\ref{fig:compare}, we introduce a searching event (e.g., ``Alice comes back and walks around"), enabling the agent to perceive both intention and false belief. Furthermore, our Level-2 Mental Reasoning is \textit{Robot-Centric}, requiring inference of both the agent’s and the human’s mental states, whereas existing benchmarks adopt a role-centric design that infers only the human’s reasoning via multiple-choice questions.

% \noindent \textbf{Construction Pipeline.}
% We use VirtualHome~\cite{puig2018virtualhome} and ThreeDWorld~\cite{gan2020threedworld} to simulate realistic household environments. The data collection process consists of three stages:
\noindent \textbf{Construction Pipeline.}
We use VirtualHome~\cite{puig2018virtualhome} and ThreeDWorld~\cite{gan2020threedworld} to simulate realistic household environments. The pipeline consists of three stages:
(1) \textbf{Story Construction.}
% Inspired by a taxonomic approach, we first generate initial story scripts based on factors such as room type, number of characters, goals, and involved objects by using GPT-4o~\cite{achiam2023gpt}. These scripts are then manually reviewed by five annotators to remove unnatural or implausible scenarios.\footnote{More details are in Supplementary Sec.B.}
We generate initial story scripts using GPT-4o~\cite{achiam2023gpt} based on room type, character setup, goals, and involved objects, followed by manual filtering by five annotators to remove implausible scenarios.\footnote{More details can be found in Sec.~B of the Supplementary Material.}
(2) \textbf{Multimodal Data Collection.}
Each script is reenacted in the simulators to collect video data. 
% In VirtualHome, humanoid agents and objects are controlled through scripting, while in ThreeDWorld, actions are performed using manual keyboard control.
% To ensure high-quality data that strictly follows the story scripts, collecting one sample takes 25–35 minutes in VirtualHome and 50–70 minutes in ThreeDWorld. Finally, we collect 590 examples. 
Ensuring strict adherence to scripts, each sample takes 25-35 minutes in VirtualHome and 50-70 minutes in ThreeDWorld, yielding 590 examples in total.
(3) \textbf{MindPower Reasoning Hierarchy Annotation.}
 Five trained annotators label all six layers of the MindPower Reasoning Hierarchy for every sample. \footnote{Details about videos and labels are provided in Sec.~B of the Supplementary Material.}

% \noindent \textbf{Data Statistics.} Our dataset covers 8 apartment layouts and 17 character models, including children, female and male adults, and wheelchair users. Each apartment contains common functional rooms such as a bedroom, kitchen, bathroom, and living room, along with typical household objects.

\subsection{Data Statistics}
% Finally, the MindPower Benchmark contains 590 examples, each including a video and a MindPower Reasoning Hierarchy text label. Among them, 37 examples come from MuMA-ToM~\cite{shi2025muma}, for which we added the ``Stage 3" scenarios as shown in Fig.\ref{fig:compare}. Two examples come from the CHAIC~\cite{du2024constrained}.
% The average video duration is 44.05 seconds, and the average text length is 164.94 words. Among these videos, 113 feature a single humanoid agent, 373 involve two humanoid agents, and 104 include three humanoid agents.

\noindent \textbf{Task Diversity.} Our benchmark incorporates 2 simulators, 8 home layouts, and 16 humanoid agents representing different age groups, genders, and mobility conditions, including children, adults, and wheelchair users. \footnote{More examples can be found in Sec.~B of the Supplementary Material.}
% This diversity enables modeling human-scene interactions under physical constraints such as limited leg mobility or height-related accessibility challenges.  

% \noindent \textbf{Robot-cetric.} As shown in Tab.~\ref{tab:tom-datasets},
% existing ToM datasets, such as MuMA-ToM~\cite{shi2025muma} and MMToM-QA~\cite{jin2024mmtom}, focus on inferring the beliefs or intentions of story characters, with evaluations framed as multiple-choice question answering. While useful for assessing recognition of mental states, these settings do not capture realistic decision-making, and the reasoning remains confined to the viewpoint of narrative characters—corresponding only to Level-2 reasoning in our MindPower hierarchy.
% Robotic benchmarks, such as AToM-Bot~\cite{ding2024atom}, introduce embodied decision-making but are limited to reasoning about hidden beliefs, with narrow goal spaces and constrained task diversity.

% In contrast, our MindPower Benchmark bridges these gaps by integrating explicit ToM reasoning with autonomous decision and action generation. It enables reasoning from the agent’s own belief perspective and adopts an open-ended format that jointly evaluates false-belief correction and implicit goal inference—two core components of human-like social cognition.
\noindent \textbf{Robot-Centric Perspective.}
As summarized in Tab.\ref{tab:tom-datasets} and Fig.~\ref{fig:compare}, existing ToM Benchmark, such as MuMA-ToM~\cite{shi2025muma} and MMToM-QA~\cite{jin2024mmtom} primarily assess the understanding of beliefs or intentions in narrative settings, 
typically through multiple-choice question answering. 
At the Mental Reasoning level, they only infer the current human’s mental state, while MindPower can infer the mental states of both the human and the embodied agent, and additionally perform Level-3 Decision Making and Action.
% Moreover, robotic benchmarks such as AToM-Bot~\cite{ding2024atom} can make decision but still focus narrowly on hidden-belief reasoning, offering limited goal spaces and task diversity.

\medskip
\noindent
In contrast, our \textbf{MindPower Benchmark} bridges these gaps by integrating explicit Mental Reasoning with autonomous decision making and action generation.
It enables reasoning from the agent’s own perspective and adopts an \textbf{Open-Ended} format that jointly evaluates 
False-Belief Correction and Implicit Goal Inference \& Completion. We will introduce the proposed evaluation metrics in Sec.~\ref{Experiment}.

\begin{figure}[t]
  \centering

  % 上方子图
  \begin{subfigure}[t]{1.0 \linewidth}
    \centering
    \includegraphics[width=\linewidth]{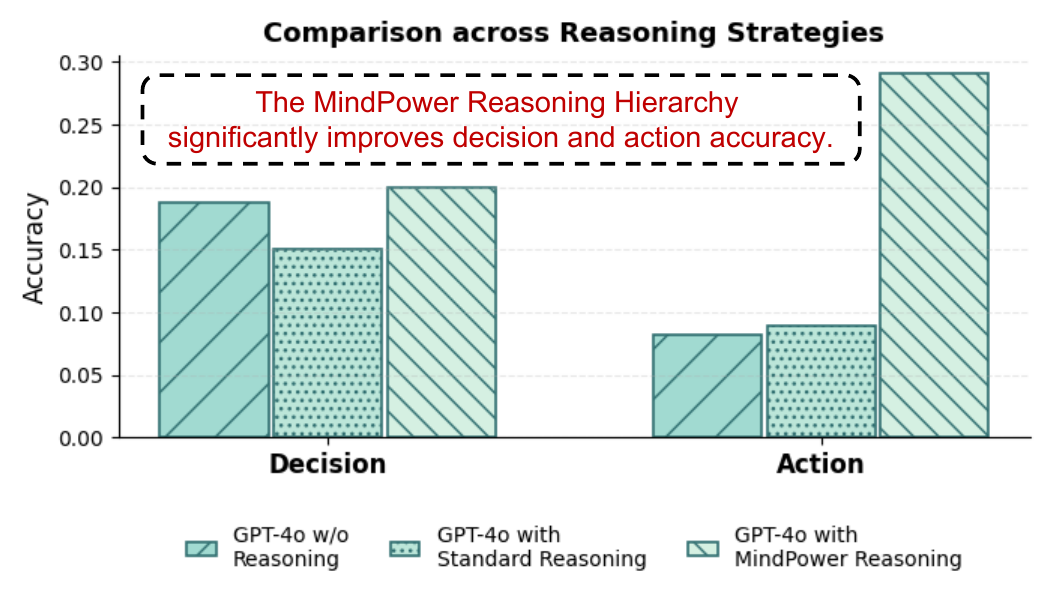}
    \caption{Experiment on different reasoning methods.}
    \label{fig:decision}
  \end{subfigure}

  \vskip 0.4em % 控制上下间距

  % 下方子图
  \begin{subfigure}[t]{1.0 \linewidth}
    \centering
    \includegraphics[width=\linewidth]{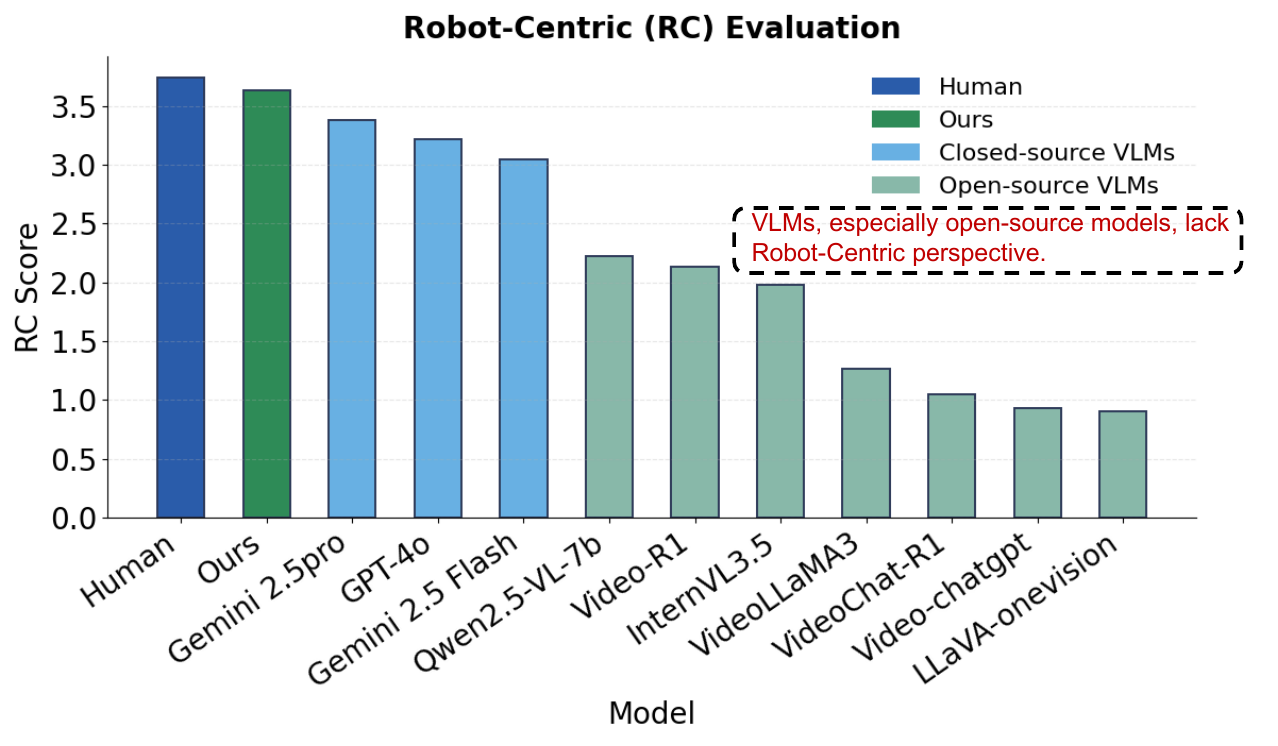}
    \caption{Model comparison on Robot-Centric (RC) score. ``Ours" is the model we propose in Sec.~\ref{sec:method}.}
    \label{fig:action}
  \end{subfigure}

  \caption{\textbf{Experiments on MindPower Benchmark.}}
  \label{fig:benchmark}
\end{figure}

\subsection{Experiment  and Discussion on MindPower Benchmark}
\label{bench3.4}

% \textbf{Conclusion front}

We split the dataset into training and testing sets with an 8:2 ratio and evaluated it on human participants as well as both open-source and closed-source Vision Language Models (VLMs). The detailed results are presented in \cref{tab:maintab}. 

We summarize our main findings as follows:

(1) \textbf{Human participants achieved the highest scores, clearly outperforming all VLMs.} 
Specifically, 9 trained participants were asked to watch the collected videos and provide BDI reasoning processes, followed by corresponding decisions and actions. As shown in bottom-right of Fig.\ref{fig:overview}, humans surpass all VLMs.

% \noindent\textbf{Closed-source VLMs achieve the best overall performance in ToM Reasoning and decision-making.}
(2) \textbf{Closed-source VLMs showed superior results in Perception, Mental Reasoning, and Decision Making and Action, with Gemini-2.5 Pro and GPT-4o achieving the highest scores.} As shown in Tab.~\ref{tab:maintab}, among open-source VLMs, those with reasoning abilities, such as Video-R1~\cite{feng2025video} and VideoChat-R1~\cite{li2025videochat}, performed the best.

(3) \textbf{The MindPower Reasoning Hierarchy substantially improves decision and action accuracy (Level-3).}  
To further validate the effectiveness of the MindPower Benchmark and the proposed MindPower Reasoning Hierarchy, we conducted ablation studies by removing Level-1 and 2 and instructing models to directly output \textit{Decision} and \textit{Action} results. We evaluated this setup on GPT-4o~\cite{achiam2023gpt}. As shown in Fig.\ref{fig:decision}, removing the MindPower Reasoning Hierarchy led to a clear performance degradation: GPT-4o's decision-making accuracy dropped by 1.24\%, while action generation accuracy decreased from 2.91\% to 0.82\%, demonstrating that the MindPower Reasoning Hierarchy is crucial for improving both the quality and consistency of decision and action outputs. Moreover, when using standard step-by-step reasoning (\texttt{<think>} ... \texttt{</think>}) instead of the MindPower Reasoning Hierarchy, performance degrades substantially: decision accuracy falls by 4.89\%, and action accuracy decreases from 2.91\% to 0.90\%. These results indicate that the MindPower Reasoning Hierarchy significantly improves the accuracy of both decision-making and action generation compared with standard reasoning.

% \noindent\textcolor{red}{\textbf{Why VLMs perform poorly?}}
% \begin{itemize}
%     \item \textit{Inconsistent BDI Reasoning: accuracy decreases as reasoning levels increase.} 
%     As the task progresses from Level-1 (Multimodal Perception) to Level-2 (Mental Reasoning), the difficulty gradually increases. Results from VLMs show a clear performance drop across levels. \textcolor{red}{For example, GPT-4o's sentence score decreases from 50.94\% at the perception level to 28.20\% at the intention level. Similarly, Video-R1 drops from 50.39\% to 28.76\%.} Overall, VLMs struggle more as the reasoning level increase and hard to struggle to maintain consistency across the MindPower reasoning process.
(4) \textbf{VLMs, especially open-source models, lack a Robot-Centric Perspective.}  
During Perception level, VLMs often provide general video descriptions about clothing or the environment instead of focusing on individual actions. They overlook crucial details such as movements, directions, and appearance sequences, which are essential for inferring implicit goals or detecting false beliefs.  
At higher reasoning layers, they are easily biased by the environment rather than reasoning from a \textbf{Robot-Centric Perspective} of both human and robot mental states. For example, in a kitchen scene, a model may predict cleaning kitchenware, while the person is actually searching for an item that someone else has taken. In a bedroom, it may assume tidying the bed, even though the person is only retrieving something from it.  
Overall, from Perception to Mental Reasoning, VLMs fail to adopt a Robot-Centric Perspective and to reason from specific actions or contradictions. Instead, they rely on coarse and stereotypical descriptions.  
As shown in Fig.~\ref{fig:action}, we use GPT-4o~\footnote{Prompt can be found in Sec.~C of  the Supplementary Material.} to evaluate whether VLMs consider individual actions and contradictions in human behavior. The results show that open-source VLMs still exhibit a substantial gap compared with human reasoning.

\section{Mind-Reward for ToM Reasoning}
\label{sec:method}
After data collection, we propose our method to let VLMs learn to act from ToM Reasoning. Our method is guided by two core principles:

\begin{itemize}
    \item \textbf{BDI Consistency.} 
    The reasoning hierarchy from \texttt{<Perception>} to \texttt{<Belief>}, \texttt{<Desire>}, \texttt{<Intention>}, \texttt{<Decision>}, and \texttt{<Action>} should remain logically consistent across all layers.
    
    \item \textbf{Robot-Centric Optimality.} 
    The agent must reason and act from its own embodied perspective. 
    During the Mental Reasoning level, it simultaneously infers its own beliefs and performs second-order reasoning about the human’s beliefs, maintaining correct perspective separation.
\end{itemize}

\noindent
Following this design, we adopt a two-stage training paradigm similar to Visual-RFT~\cite{liu2025visual} and DeepSeekMath~\cite{shao2024deepseekmath}. 
Specifically, we first perform Supervised Fine-Tuning (SFT) to establish base reasoning alignment, followed by Group Relative Policy Optimization (GRPO) using our proposed reward, combining \textbf{Mind-Reward} and \textbf{Format-Reward}, to enhance BDI consistency and Robot-Centric optimality.

\begin{figure}[t]
  \centering
  \includegraphics[width=1\linewidth]{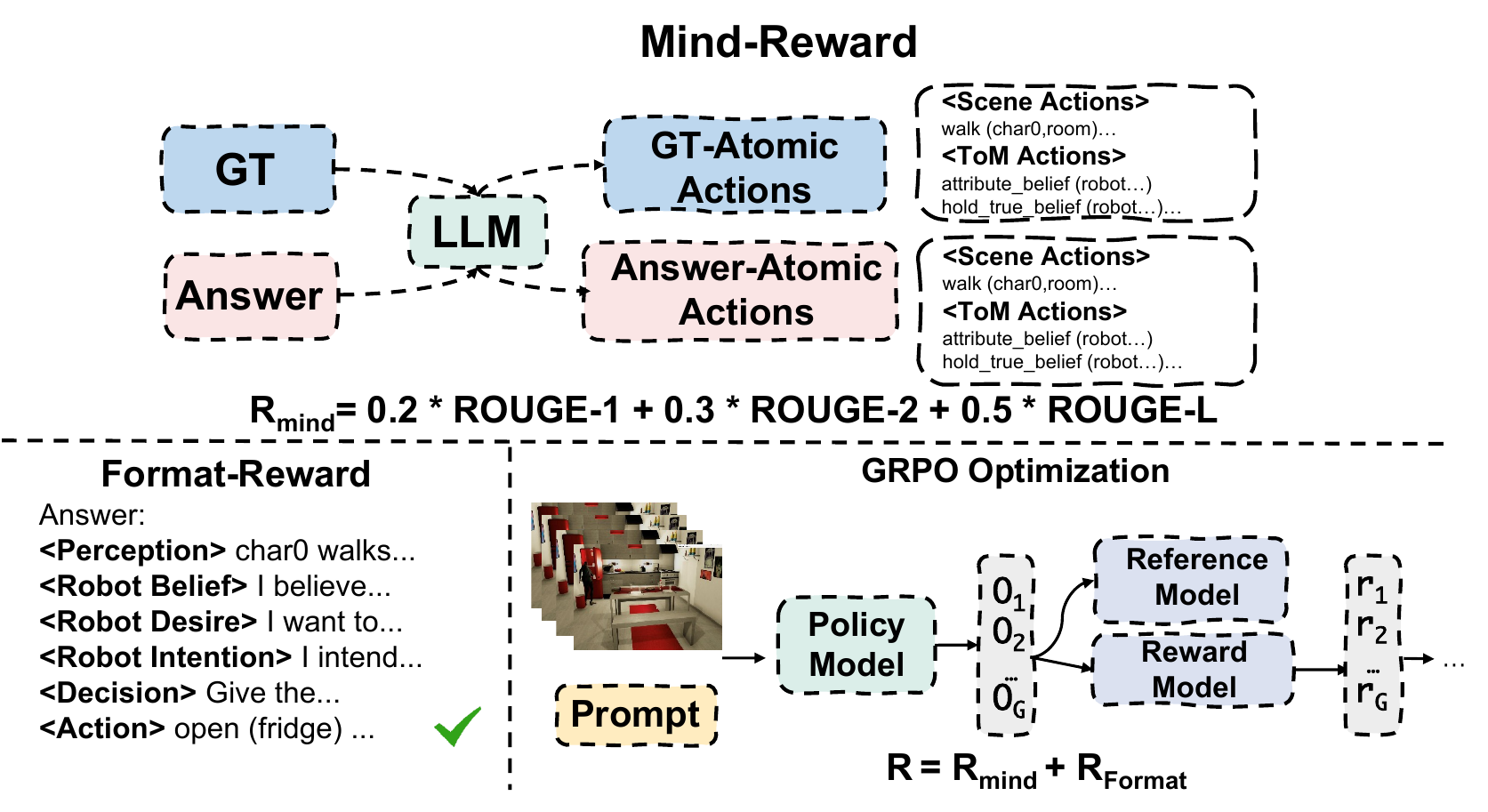} 
  % \caption{\textbf{ The reward we use in GRPO.} It combines Mind-Reward and Format-Reward.}
  \caption{\textbf{Reward Formulation.} The overall reward integrates both the Mind-Reward and the Format-Reward components.}
  \label{fig:method}
\end{figure}

\noindent \textbf{Mind-Reward.} 
In the GRPO stage, we introduce Mind-Reward $R_{\text{Mind}}$ to further optimize the SFT model. The Mind Reasoning Hierarchy is continuous and requires maintaining consistency across all reasoning levels and layers. Moreover, across different reasoning layers, such as the perception of events and the inference of beliefs about embodied agents and humans, there exist inherent temporal and logical dependencies that must be preserved. \footnote{A detailed discussion is provided in Sec.~D of the Supplementary Material.}

% Traditional reward functions based on token-overlap metrics (e.g., ROUGE~\cite{lin2004rouge}), as used in Video-R1~\cite{feng2025video}, or those relying on LLM-as-judge evaluation~\cite{li2025videochat}, mainly capture surface-level lexical similarity. However, these methods fail to enforce both cross-level consistency and intra-level logical coherence.
% To address these limitations, Mind-Reward is designed to explicitly encourage:  
% (1) Perceptual Grounding,  
% (2) BDI Consistency,  
% (3) Accurate Action Generation, and  
% (4) Robot-Centric Perspective alignment.

We represent each reasoning layer (from \texttt{<Perception> to \texttt{<Action>}}) as a sequence of \textit{atomic actions}, denoted as \texttt{action (agent, object)}, where \texttt{agent} refers to the owner of the action or mental state, and \texttt{object} denotes the target entity or mental content being acted upon. Since layers in the Mental Reasoning level involves distinct cognitive and physical reasoning patterns, we construct a unified atomic-action table that encompasses both categories. \footnote{Details can be found in Sec.~D of the Supplementary Material.} Both the ground-truth and generated outputs are then converted into structured atomic action sequences by an LLM (Qwen3-Max~\cite{yang2025qwen3}) during the GRPO training process, and these extracted atomic actions are subsequently used for reward computation.

Mind-Reward evaluates reasoning quality from three complementary aspects:  
(1) \textbf{Atomic Accuracy}: measured by ROUGE-1, it quantifies the proportion of correctly matched atomic actions, each tagged with a perspective attribute (human or embodied agent) to ensure Robot-Centric Perspective alignment;
(2) \textbf{Local Consistency}: measured by ROUGE-2 between adjacent atomic pairs to assess short-range reasoning coherence;
(3) \textbf{Global Consistency}: measured by ROUGE-L (longest common subsequence) to evaluate the overall reasoning alignment across the reasoning process.

The final reward is a weighted sum of these components:
\begin{equation}
R_{\text{Mind}} = \alpha_1 R_{\text{atomic}} + \alpha_2 R_{\text{local}} + \alpha_3 R_{\text{global}}.
\end{equation}

This reward formulation explicitly enforces both ToM consistency and Robot-Centric Perspective throughout GRPO.

% \noindent \textbf{Stage 1: Mind Reasoning Supervised Finetuning.}

% \noindent \textbf{Stage 2: Mind Reasoning Reinforcement Finetuning.}

\noindent \textbf{Format-Reward.} 
Format-Reward $R_{\text{Format}}$ is computed by performing a sequential regular expression match over the six reasoning layers: 
\texttt{<Perception>}, \texttt{<Belief>}, \texttt{<Desire>}, \texttt{<Intention>}, \texttt{<Decision>}, and \texttt{<Action>}. 
If all layers appear in the correct order, the reward is set to 1; otherwise, it is 0.

\noindent \textbf{Overall Reward.} 
As shown in Fig.~\ref{fig:method}, the final reward $R$ used in GRPO combines the proposed Mind-Reward $R_{\text{Mind}}$ and Format-Reward $R_{\text{Format}}$ as:
\begin{equation}
R = R_{\text{Mind}} + R_{\text{Format}}.
\end{equation}

The advantage $A_i$ is then computed within each group as:
\begin{equation}
A_i = \frac{R_i - \text{mean}(\{R_j\})}{\text{std}(\{R_j\})},
\end{equation}
where $R_i$ is the reward for the $i$-th response.

\noindent \textbf{Optimization.}
We adopt the GRPO algorithm proposed in DeepSeekMath~\cite{shao2024deepseekmath} to optimize the model.
GRPO samples a group of outputs $\{o_1, o_2, \cdots, o_G\}$ from the old policy $\pi_{\theta_{\text{old}}}$ and updates the policy $\pi_{\theta}$ by maximizing the following objective:

\begin{multline}
J_{\text{GRPO}}(\theta) = 
\mathbb{E}_{q \sim P(Q), \{o_i\}_{i=1}^{G} \sim \pi_{\theta_{\text{old}}}(O|q)} \\
\Bigg[
\frac{1}{G} \sum_{i=1}^{G} 
\min \Bigg(
\frac{\pi_{\theta}(o_i|q)}{\pi_{\theta_{\text{old}}}(o_i|q)} A_i, \;\\
\text{clip}\!\left(
\frac{\pi_{\theta}(o_i|q)}{\pi_{\theta_{\text{old}}}(o_i|q)}, 
1 - \epsilon, 1 + \epsilon
\right) A_i
\Bigg) 
- \beta D_{\text{KL}}\!\left(\pi_{\theta} \| \pi_{\text{ref}}\right)
\Bigg].
\end{multline}

\begin{table*}[htbp]
\centering
\caption{\textbf{Quantitative Evaluation.} We evaluate our model against both image-based and video-based VLMs. “B” denotes the BERTScore, “S” represents the Sentence Transformer score, and “BPC” means BDI and Perspective Consistency. The BPC score ranges from 0 to 10, while all other metrics are normalized to a range of 0 to 100.}
\label{tab:maintab}
\resizebox{\textwidth}{!}{%
\begin{tabular}{l c c c c c c c c c c c c c}
\toprule
\multirow{2}{*}{\textbf{Method}} 
  & \multicolumn{2}{c}{\textbf{Perception}} 
  & \multicolumn{2}{c}{\textbf{Belief}} 
  & \multicolumn{2}{c}{\textbf{Desire}} 
  & \multicolumn{2}{c}{\textbf{Intention}} 
  & \multicolumn{2}{c}{\textbf{Decision}} 
  & \multicolumn{2}{c}{\textbf{Action}} 
  & \multirow{2}{*}{\shortstack{\textbf{BPC}}} \\
\cmidrule(lr){2-3}\cmidrule(lr){4-5}\cmidrule(lr){6-7}\cmidrule(lr){8-9}\cmidrule(lr){10-11}\cmidrule(lr){12-13}
 & B & S & B & S & B & S & B & S & B & S & SR & AC & \\
\midrule

% ---------------- Human Study ----------------
% \rowcolor{gray!10}
\multicolumn{14}{l}{\textit{Human Study}} \\
% \midrule
Human Baseline
  & - & - & 47.65 & 61.81 & 46.76 & 53.71 & 39.18 & 52.93 & 34.55 & 56.66 & 19.37 & 26.26 & 8.19 \\
\hdashline

% ---------------- Video-input ----------------
% \rowcolor{gray!10}
\multicolumn{14}{l}{\textit{Video-input}} \\
% \midrule
Gemini-2.5 Flash~\cite{comanici2025gemini}
  & 31.10 & 48.36 & 29.07 & 38.64 & 28.36 & 30.69 & 19.05 & 29.04 & 21.68 & 34.57 & 1.38 & 1.35 & 8.72 \\
Gemini-2.5 Pro~\cite{comanici2025gemini}
  & 24.62 & 43.43 & 32.02 & 36.79 & 31.38 & 30.21 & 22.65 & 30.33 & 24.23 & 33.87 & 2.08 & 2.54 & 8.56 \\
Qwen2.5-VL-7B-Instruct~\cite{bai2025qwen2}
  & 26.05 & 38.20 & 20.27 & 28.43 & 26.05 & 22.93 & 16.01 & 23.21 & 16.69 & 26.56 & 0.29 & 0.22 & 6.07 \\
VideoLLaMA3-7B~\cite{zhang2025videollama}
  & 14.80 & 31.86 & 7.82 & 30.08 & 8.09 & 21.76 & 4.61 & 24.28 & 5.34 & 19.59 & 0.63 & 0.60 &  5.33 \\
InternVL3.5-8B~\cite{wang2025internvl3}
  & 23.23 & 42.26 & 21.98 & 26.90 & 22.20 & 22.45 & 16.53 & 23.21 & 15.64 & 28.76 & 0.10 & 0.08 & 6.52 \\
Video-LLaVA~\cite{lin2023video}
  & 2.96 & 25.33 & 5.05 & 14.87 & 6.82 & 15.55 & 16.63 & 15.30 & 3.29 & 19.50 & 0.08 & 0.08 & 4.81 \\
Video-ChatGPT~\cite{maaz2023video}
  & 7.04 & 27.00 & 9.90 & 25.72 & 5.16 & 16.79 & 2.70 & 21.44 & 1.46 & 19.95 & 0.00 & 0.00 & 5.52 \\
VideoChat-R1~\cite{li2025videochat}
  & 27.47 & 42.47 & 21.57 & 30.11 & 22.56 & 20.36 & 15.03 & 24.70 & 17.21 & 25.71 & 0.64 & 0.82 &  6.00 \\
Video-R1~\cite{feng2025video}
  & 30.56 & 47.46 & 25.56 & 34.58 & 26.68 & 29.17 & 17.13 & 27.56 & 18.91 & 30.33 & 1.43 & 1.72 & 6.45 \\
\hdashline

% ---------------- Image-input ----------------
% \rowcolor{gray!10}
\multicolumn{14}{l}{\textit{Image-input}} \\
% \midrule
GPT-4o~\cite{achiam2023gpt}
  & 33.07 & 48.37 & 30.05 & 39.47 & 31.16 & 32.75 & 16.16 & 29.55 & 19.96 & 34.35 & 1.82 & 2.91 & 8.05 \\
Qwen2.5-VL-7B-Instruct~\cite{bai2025qwen2}
  & 24.89 & 39.97 & 19.46 & 29.21 & 22.59 & 19.14 & 16.80 & 23.49 & 19.11 & 23.79 & 0.15 & 0.15 & 6.72 \\
InternVL3.5-8B~\cite{wang2025internvl3}
  &6.43 & 18.78 & 15.71&
20.77&
19.30&
17.38&
13.97&
19.72& 12.62 & 18.77 & 0.00 & 0.00 & 5.95 \\
LLaVA-OV-8B~\cite{li2024llava}
  & 8.08 & 26.45 & 15.09 & 23.21 & 22.31 & 21.40 & 16.21 & 19.58 & 17.11 & 21.25 & 0.00 & 0.00 & 6.45 \\
\hdashline
  % \rowcolor{gray!10}
\multicolumn{14}{l}{\textit{Ours}} \\
% \midrule
% \rowcolor{softlavender}
% \rowcolor{softblue}
\rowcolor{tablegray}

% \textbf{Ours-SFT}
%   & 46.23 & 60.78 & 51.55 & 47.51 & 53.66 & 47.00 & 40.32 & 44.48 & 43.68 & 48.93 & 14.20 & 188.10 & 8.08 \\
% \rowcolor{softlavender}
Mind-Reward only
  & 21.84 & 39.99 & 18.70 & 27.81 & 21.35 & 18.85 & 21.90 & 23.30 & 17.58 & 24.68 & 0.28 & 0.40 & 6.63 \\
% \rowcolor{softlavender}
% \rowcolor{softblue}
\rowcolor{tablegray}
  SFT only
  & 32.78 & 52.72 & 43.15 & 42.48 & 47.01 & 37.83 & 34.86 & 39.48 & 36.70 & 43.84 & 8.50 & 10.48 & 8.78 \\
% \rowcolor{softlavender}
%   No atomic extraction
%   & & & & &  &  &  &  &  &  & &  &  \\
  
% \rowcolor{softlavender}
% \rowcolor{softblue}
\rowcolor{tablegray}

% \textbf{Ours-SFT}
%   & 46.23 & 60.78 & 51.55 & 47.51 & 53.66 & 47.00 & 40.32 & 44.48 & 43.68 & 48.93 & 14.20 & 188.10 & 8.08 \\
% \rowcolor{softlavender}
\textbf{Ours (SFT+Mind-Reward)}
  & \textbf{44.79} & \textbf{59.93} & \textbf{49.14} & \textbf{46.49} & \textbf{51.25} & \textbf{45.75} & \textbf{37.79} & \textbf{42.57} & \textbf{40.17} & \textbf{47.12} & \textbf{11.75} & \textbf{15.40} & \textbf{8.87} \\
\bottomrule
\end{tabular}%
}
\end{table*}

\section{Experiment}
\label{Experiment}
\subsection{Evaluation Metrics}

We design evaluation metrics to assess the model’s performance across three levels, corresponding to the full reasoning hierarchy from \texttt{<Perception>} to \texttt{<Decision>} and \texttt{<Action>}.

\noindent \textbf{Level-1: Perception.} The perception module outputs textual descriptions (captions). We evaluate these outputs using \textit{BERTScore}~\cite{zhang2019bertscore} and \textit{Sentence Transformer}~\cite{reimers-2020-multilingual-sentence-bert} similarity, which measure the semantic alignment between the generated captions and the ground-truth descriptions.

\noindent \textbf{Level-2: Mental Reasoning.} We similarly evaluate the reasoning outputs using \textit{BERTScore} and \textit{Sentence Transformer} similarity, measuring semantic consistency across the three components of \texttt{<Belief>}, \texttt{<Desire>}, and \texttt{<Intention>}.

\noindent \textbf{Level-3: Decision Making and Action.} The decision stage generates textual outputs, which are evaluated using the same \textit{BERTScore} and \textit{Sentence Transformer} similarity metrics. The action stage produces sequences of atomic actions, which are evaluated using two additional metrics: \textit{Success Rate (SR)} and \textit{Action Correctness (AC)}. These metrics assess both the overall correctness of the action sequence and the accuracy of each atomic action, represented in the form \texttt{action (object)}. The \textit{SR} score combines multiple ROUGE components and is defined as:

\begin{equation}
\text{SR} = \frac{2R_{1} + 3R_{2} + 5R_{L}}{10},
\end{equation}
where \( R_{1} \), \( R_{2} \), and \( R_{L} \) denote the ROUGE-1, ROUGE-2, and ROUGE-L scores, respectively. 
The \textit{AC} score measures how accurately the generated action sequence \( A^{*} \) matches the ground-truth sequence \( \hat{A} \), and is computed as:
\begin{equation}
\text{AC} = \left\lfloor \frac{|A^{*} \cap \hat{A}|}{|\hat{A}|} \right\rfloor,
\end{equation}
where $|A^{*} \cap \hat{A}|$ denotes the number of atomic actions in $A^{*}$ that correctly match the ground-truth sequence $\hat{A}$, and $|\hat{A}|$ is the total number of actions in the ground-truth sequence.

\begin{figure*}[t]
  \centering
  \includegraphics[width=1\linewidth]{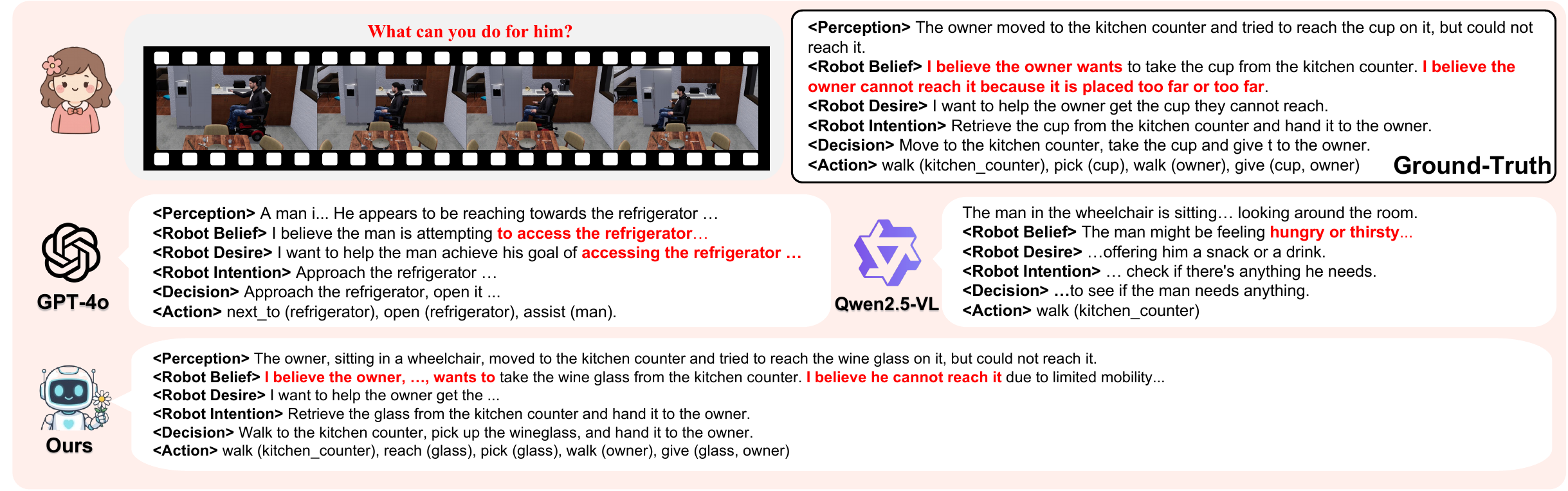} 
  \caption{\textbf{Qualitative Evaluation.} We compare our model with GPT-4o and Qwen2.5-VL-7B-Instruct. Although GPT-4o outputs the correct format, it incorrectly infers that the human intends to open the refrigerator. In contrast, Qwen2.5-VL-7B-Instruct fails to follow the required format and also produces incorrect mental reasoning. Detailed outputs are provided in Sec.~D of the Supplementary Material. }
  \label{fig:vis}
\end{figure*}

\noindent \textbf{BDI and Perspective Consistency.}
We use GPT-4o to evaluate the BDI consistency and perspective of the generated outputs. The content from \texttt{<Perception>} to \texttt{<Action>} is assessed by GPT-4o based on three criteria: (1) whether each reasoning layer logically follows from the previous one without contradictions, (2) whether the overall reasoning is complete and precise, and (3) whether the reasoning genuinely adopts the robot’s perspective and effectively assists the human characters in the story.

\subsection{Experiment Setup}

We randomly split the dataset into training and testing sets with an 8:2 ratio. We used Qwen2.5-VL-7B-Instruct as the base model. We extracted 32 frames from each video and concatenated them for training. We used 5 training epochs for SFT and 400 iterations for GRPO. The number of generations was set to 8, and training was done on a single H800 GPU. We set $\alpha_1$ as 0.2, $\alpha_2$ as 0.3, and $\alpha_3$ as 0.5.

\subsection{Quantitative Evaluation}
We evaluated several closed-source baselines, including Gemini-2.5 Pro~\cite{comanici2025gemini}, Gemini-2.5 Flash~\cite{comanici2025gemini}, and GPT-4o~\cite{achiam2023gpt}. Since GPT-4o does not accept raw video input, we uniformly sampled an average of 64 frames as its input. For open-source baselines, we tested Qwen2.5-VL-7B-Instruct~\cite{bai2025qwen2}, InternVL3.5-8B~\cite{wang2025internvl3}, Video-LLaVA3~\cite{lin2023video}, Video-ChatGPT~\cite{maaz2023video}, Video-R1~\cite{feng2025video}, VideoChat-R1~\cite{li2025videochat}, and LLaVA-OV-8B~\cite{li2024llava}.
For Qwen2.5-VL-7B-Instruct and InternVL3.5-8B, we conducted evaluations under two settings: (1) frame-averaged input and (2) direct video input.
Experimental results demonstrate that our model achieves the best overall performance in \textbf{Perception}, \textbf{Mental Reasoning}, and \textbf{Decision Making and Action} levels. As shown in Tab.~\ref{tab:maintab}, 
compared with the Qwen2.5-VL-7B-Instruct, our model achieves a +20.04\% improvement in \textit{Sentence Transformer} score for perception and a +23.33\% gain in the \texttt{<Decision>} layer. Moreover, the \textit{SR} increases by 11.6\%, and the \textit{AC} improves by 15.25\%.
Notably, InternVL3.5-8B, LLaVA-OV-8B, and Video-ChatGPT obtain zero scores on the \textit{SR} and \textit{AC}, as their outputs mainly consist of non-executable expressions such as \texttt{identify ()} or \texttt{scan ()}, rather than concrete, goal-directed action commands.

\noindent\textbf{Ablation Study.}
% We conducted the following experiments: (1)The model only using  SFT;
% (2)The model without SFT, using only Mind-Reward;
% (3) A model in which the step for extracting atomic operations in Mind-Reward is removed, and the reward is computed directly. In this case, we calculate a combination of different ROUGE scores between the ground-truth and generated outputs, which is the same as Video-R1~\cite{feng2025video}.
As shown in Tab.~\ref{tab:maintab}, using only SFT yields a certain improvement, indicating that the MindPower Reasoning Hierarchy enhances the model's mental reasoning and decision-making capabilities. Further improvements are observed when incorporating Mind-Reward, demonstrating that it can further strengthen the model’s performance.
Without SFT, we find that compared with the initial Qwen2.5-VL-7B-Instruct, although there is some improvement in decision and action accuracy, the overall performance remains suboptimal. This indicates that the model still requires SFT for effective cold-start training.

\subsection{Qualitative Evaluation}
% We take the example of a man in a wheelchair trying to reach a distant cup to compare GPT-4o and Qwen2.5-VL-7B-Instruct in Fig.~\ref{fig:compare}. 
We illustrate the differences between GPT-4o and Qwen2.5-VL-7B-Instruct using a scenario where a man in a wheelchair attempts to reach a distant cup (Fig.~\ref{fig:compare}).
% We find that both models are affected by environmental cues: GPT-4o misinterprets the scene as opening a refrigerator, while Qwen2.5-VL-7B-Instruct assumes the human is hungry. 
Both models are easily swayed by environmental cues: GPT-4o hallucinates a refrigerator-opening action, whereas Qwen2.5-VL-7B-Instruct infers hunger.
As discussed in Sec.~\ref{bench3.4}, these failures arise from the lack of Robot-Centric Perception. In contrast, our model infers the human’s inability to reach the cup and performs second-order reasoning by clearly separating perspectives.

\section{Conclusion and Future Work}

In this work, we introduce the MindPower Benchmark, which incorporates the Robot-Centric MindPower Reasoning Hierarchy with three levels and six layers for modeling ToM Reasoning. The benchmark includes the MindPower Dataset with two tasks, False-Belief Correction and Implicit Goal Inference \& Completion, together with evaluation metrics for assessing whether VLM-based embodied agents can perform decision making and action generation grounded in ToM Reasoning. Finally, we evaluate a variety of VLMs on our benchmark and propose a Mind-Reward mechanism that achieves the best overall performance.

In future work, we will extend the MindPower Reasoning Hierarchy to human–robot collaboration and multi-agent coordination, and deploy our model on real robots to assess its performance in practical settings.
\clearpage
\clearpage
\clearpage
\appendix
\setcounter{page}{1}
\maketitlesupplementary

Considering the space limitations of the main paper, we provide additional results and discussions in this appendix. The appendix is organized to first clarify the \textbf{key concepts} used throughout the paper, followed by detailed descriptions of our \textbf{dataset collection and annotation process}, comparisons with other benchmarks, and the prompts used in Sec.~3.4. We then describe how textual instructions are converted into atomic action sequences in the \textbf{Mind-Reward} framework. Next, we present additional experimental results, including \textbf{evaluation metrics} and \textbf{task-specific experiments}. We further discuss \textbf{potential extensions of our dataset}, such as multi-view extension and its connection to low-level execution models. Finally, we summarize the \textbf{limitations of the current benchmark and future directions for improvement}. \textbf{The full benchmark will be publicly released to encourage future research.}

\begin{enumerate}[label=\Alph*.]  % 使用大写字母编号
    \item\hyperref[sec:def]{\textbf{Definition of Terms}}
    
    \item \hyperref[sec:mindpower]{\textbf{More Details of MindPower Benchmark}}  
    \begin{enumerate}[label=\arabic*.] % 子列表保持数字编号
        \item \hyperref[sec:mindpower:story]{Details of Story Construction and Data Annotation}
        \item \hyperref[sec:mindpower:cmp]{Comparison with Other Benchmarks}
        \item \hyperref[sec:mindpower:sim]{Simulators}
        \item \hyperref[sec:mindpower:examples]{Detailed Examples of Fig.~1 and 3 in the Manuscript}
        \item \hyperref[sec:mindpower:exp]{Details of Experiments on Different Reasoning Methods}
        \item \hyperref[sec:mindpower:robot]{Robot-Centric Scoring}
        
    \end{enumerate}

    \item \hyperref[sec:mindreward]{\textbf{More Details of Mind-Reward}}  
    \begin{enumerate}[label=\arabic*.]
        \item \hyperref[sec:mindreward:atomic]{Atomic Action Table}
        \item \hyperref[sec:mindreward:discussion]{Discussion}

    \end{enumerate}

    \item \hyperref[sec:results]{\textbf{Additional Experimental Results}}  
    \begin{enumerate}[label=\arabic*.]
        \item \hyperref[sec:results:metrics]{Details of Metrics}
        \item \hyperref[sec:results:exp]{Experiments on False-Belief Correction and Implicit Goal Inference \& Completion}
        \item \hyperref[sec:results:vis]{Detailed Example of Fig.~6 in the Manuscript}
        % \item Qualitative Evaluation
    \end{enumerate}

    \item \hyperref[sec:ext]{\textbf{Extensions of Our Work}}  
    \begin{enumerate}[label=\arabic*.]
        \item \hyperref[sec:ext:view]{Multi-View of MindPower}
        \item \hyperref[sec:ext:lowlvl]{Relationship with Low-Level Execution Models}
        \item \hyperref[sec:ext:limitation]{Limitations and Future Work}
    
    \end{enumerate}
    \item \hyperref[sec:demo]{\textbf{Demo Videos}}
    
\end{enumerate}

\section{Definition of Terms}\label{sec:def}
\label{definition}

\textbf{Theory  of Mind (ToM).}
Theory of Mind (ToM)~\cite{rao1995bdi,onishi200515} is the cognitive ability to infer others’ mental states such as beliefs, desires, and intentions, and to use these inferences to predict and guide actions. ToM goes beyond perceiving observable behaviors and instead requires reasoning about what different agents know, think, and want. Higher-order ToM, including reasoning about others’ beliefs about others, is essential for coherent decision-making in multi-agent interactions involving cooperation, conflict, or deception.

\noindent \textbf{ToM Reasoning.} In our work, ``ToM Reasoning" refers to an agent’s ability to infer others’ mental states and make decisions based on them rather than solely on observable states.

\noindent \textbf{Robot-Centric.} In our work, by ``Robot-Centric" we mean that the embodied agent should reason from its own perspective. It not only needs to infer its own mental states but also reason about how it perceives the mental states of human.

\noindent \textbf{Role-Centric.} ``Role-Centric" refers to the model reasoning about mental states from the perspective of a character within the current story or multimodal input.

\noindent \textbf{MindPower Reasoning Hierarchy.} In this work, we propose that the model follows the reasoning path 
\texttt{<Perception>} $\rightarrow$ 
\texttt{<Belief>}
$\rightarrow$ 
\texttt{<Desire>}
$\rightarrow$ 
\texttt{<Intention>} $\rightarrow$ 
\texttt{<Decision>} $\rightarrow$ 
\texttt{<Action>}, 
which constitutes the MindPower Reasoning Hierarchy.

\section{More Details of MindPower Benchmark}\label{sec:mindpower}
\subsection{Details of Story Construction and Data Annotation}\label{sec:mindpower:story}
\begin{figure}[t]
  \centering
  \includegraphics[width=1\linewidth]{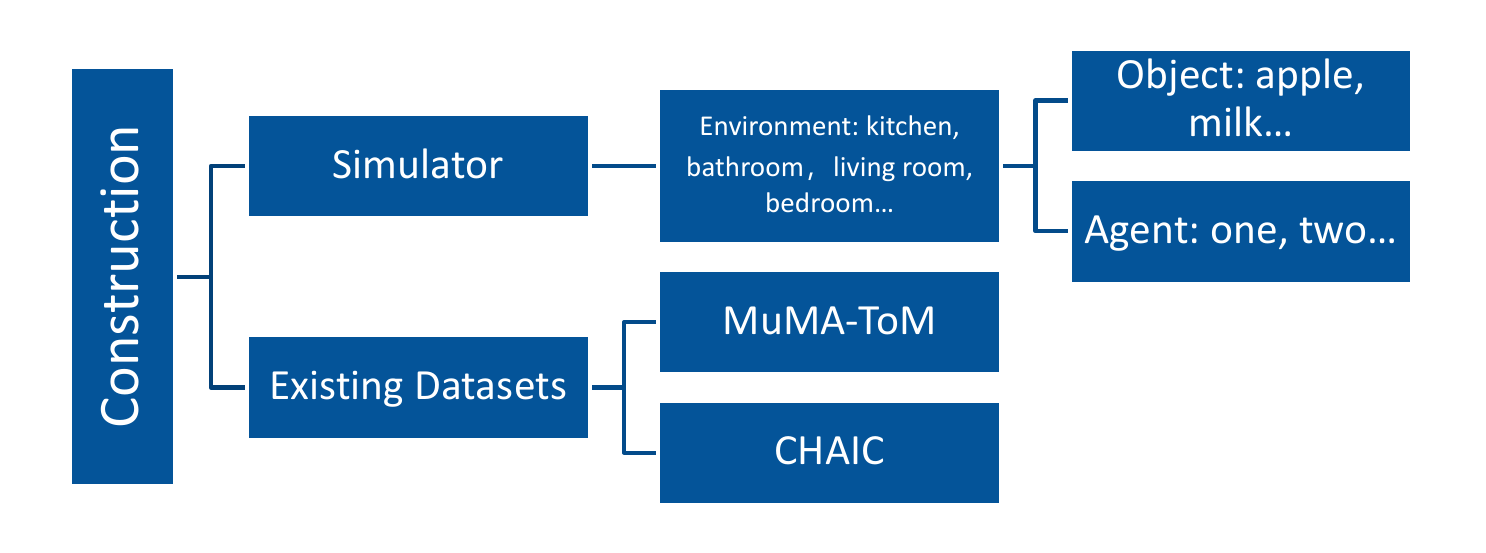} 
  \caption{Story Construction Pipeline for False-Belief Correction Task.}
  \label{fig:cons}
\end{figure}
For the \textbf{False-Belief Correction task}, as illustrated in Fig.~\ref{fig:cons}, we follow a taxonomy-driven approach. We first categorize scenarios based on the mapping between VirtualHome~\cite{puig2018virtualhome} and ThreeDWorld~\cite{gan2020threedworld} environments and the typical object distributions in each room (e.g., kitchen, living room). We then determine the number of humans involved in each scene. To cover different numbers of humanoid agents and different target (final) humanoid agents, we design three distinct prompt templates for GPT-4o to generate story scripts. When issuing each request, we iterate over a predefined list of objects along with their corresponding start and end locations. The prompts are shown in Fig.~\ref{fig:stroy_gen}.

For the \textbf{Implicit Goal Inference \& Completion task}, we design four types of scenarios to comprehensively evaluate agents’ goal-inference abilities:

(1) \textbf{Special populations.} We include scenarios featuring individuals with unique physical conditions: a wheelchair user and a 1.2-meter-tall child. A wheelchair user faces mobility and height limitations, while the child cannot reach high places. We design stories that incorporate these constraints so that the hidden goal must be inferred through contextual cues rather than physical actions.

(2) \textbf{Object-centric property reasoning.} We exploit special physical properties of household objects to construct implicit goals. For instance, since faucets can leak water, we create situations where a person leaves without turning off the faucet. Similarly, because candles provide light, we design scenes where a person reading a book suddenly experiences a power outage and begins walking around; the agent can infer that they are searching for candles (no flashlight is available in the environment).

(3) \textbf{Functional object combinations.} Based on the objects present in \textit{VirtualHome} and \textit{ThreeDWorld}, we identify typical usage pairs or triplets. For example, a knife, cutting board, and carrot together imply the goal of \textit{cutting carrots}. If a person places a cutting board on the table and puts a carrot on it before searching for another object, the hidden goal is most likely to find a knife to complete the task.

(4) \textbf{Dialogue-driven inference.} We additionally design conversational scenarios  like MuMA-ToM~\cite{shi2025muma} and FanToM~\cite{kim2023fantom} in which implicit goals must be inferred from incomplete verbal exchanges rather than direct physical interactions.

% We provide some examples: 

% \begin{tcolorbox}[colback=gray!5,colframe=gray!30,boxrule=0.3pt,left=2pt,right=2pt,top=2pt,bottom=2pt]
% \small
% \textbf{Example 1 (False-Belief Correction):}
% Char0 walked into the kitchen and picked up the cupcake. She placed the cupcake in the fridge.
% Later, Char1 walked into the kitchen, opened the fridge, picked up the cupcake, and placed it on the kitchen table.
% Then, Char2 walked into the kitchen,  picked up the cupcake, and placed it in the kitchen cabinet.
% Finally, Char0 walked back to the fridge, opened it, closed it, and walked around the kitchen, passing by the kitchen table and returning to the fridge.

% \textbf{Example 2 (Implicit Goal Inference \& Completion):}
% The man walks into the kitchen, picks up a cutting board and places it on the counter. Then he picks up a salmon and places it on the cutting board, and finally starts walking around the kitchen.

% \textbf{Example 3 (Implicit Goal Inference \& Completion for Special-Needs):}  
% A man in a wheelchair moves from the living room toward the bedroom.  
% An obstacle (e.g., a fire hydrant) blocks his way, causing him to move left, reverse, and then try again from the right.
% \end{tcolorbox}
\begin{figure*}[t]
  \centering
  \includegraphics[width=1\linewidth]{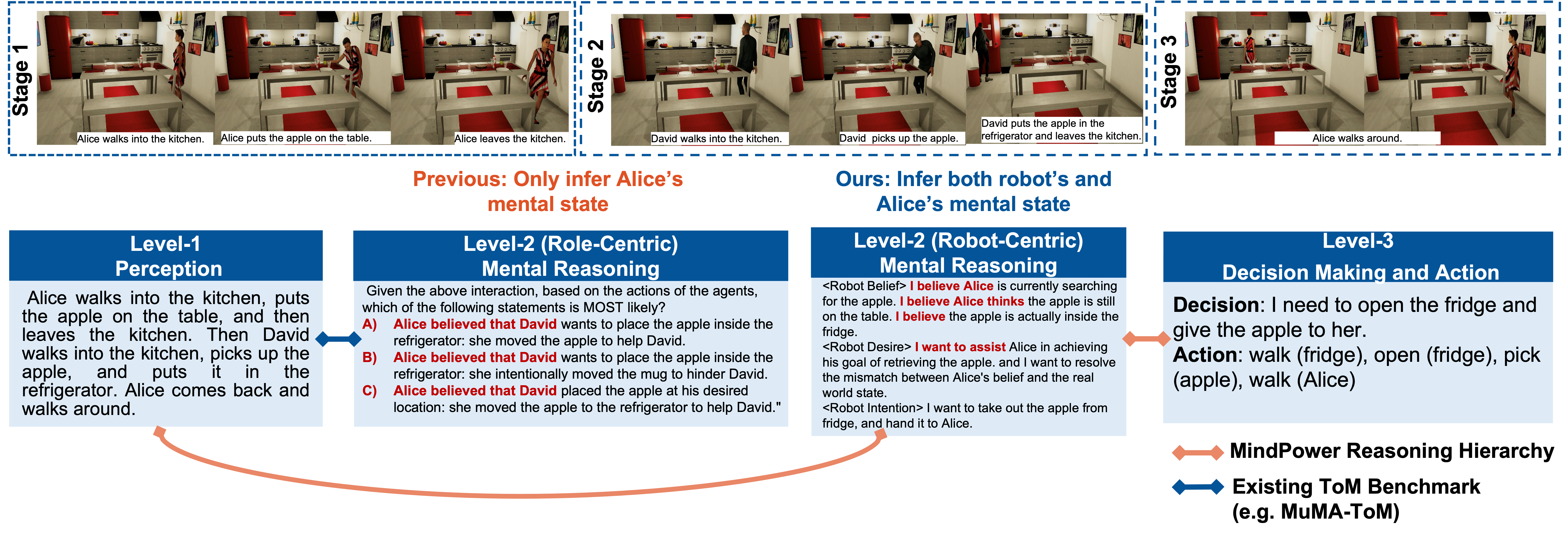} 
  \caption{Full Version of Fig.~3 in Manuscript.}
  \label{fig:full_compare}
\end{figure*}

Finally, we collect 200 examples for Implicit Goal Inference \& Completion and 390 examples for False-Belief Correction. Among them, 37 examples are adapted from MuMA-ToM~\cite{shi2025muma}, where we further augment each story by incorporating a stage-3 ``character search" segment, as illustrated in Fig.~\ref{fig:full_compare}, and 2 examples are sourced from CHAIC~\cite{du2024constrained}. Overall, 113 examples contain a single humanoid agent, 373 contain two agents, and 104 contain three agents. In addition, 17 examples involve agents with special needs, 96 focus on object-centric property reasoning and functional object combinations, and 87 correspond to dialogue-driven inference.

\textbf{Data Annotation.}
For each example in the MindPower Reasoning Hierarchy, the annotations are manually created and subsequently verified using GPT-4o~\cite{achiam2023gpt}. During the annotation process, particularly for the \texttt{<Action>} layer, we adopt a unified action space that integrates action definitions from both VirtualHome and ThreeDWorld. This approach enables us to standardize heterogeneous simulators under a single executable schema. The complete list of supported high-level actions is as follows:

\begin{tcolorbox}[colback=gray!10,colframe=black!80,title=High-Level Action Set,boxsep=2mm]
\texttt{Walk, Run, WalkTowards, WalkForward, TurnLeft, Sit, StandUp, TurnRight, Sit, StandUp, Grab, Open, Close, Put, PutIn, SwitchOn, SwitchOff, Drink, Touch, LookAt, TurnBy, TurnTo, MoveBy, MoveTo, ReachFor, ResetArm, Drop, Animate, RotateHead, ResetHead}
\end{tcolorbox}

For some examples in the False-Belief Correction task, the camera viewpoint prevents certain objects from being visible after they are moved. For instance, we design scenarios where a humanoid agent moves an object from the fridge in the kitchen to the bedroom, but the camera is fixed in the kitchen and cannot capture the final location. As a result, the embodied agent can only infer that the object has been moved, without knowing where it ends up.
In such cases, the annotated \texttt{<Action>} does not require the agent to find the object. Instead, the action is defined as reminding the returning character that the object has already been moved, thereby correcting their false belief even though the agent cannot locate the object.

\subsection{Comparison with Other Benchmarks}\label{sec:mindpower:cmp}
We compare our dataset with existing multimodal ToM benchmarks from three perspectives:

\begin{itemize}
    \item \textbf{Data source and diversity.} To the best of our knowledge, our benchmark is the first to be constructed using \textbf{two different simulators}, which substantially increases the diversity of environments, interaction patterns, and embodied tasks. In contrast, prior multimodal ToM datasets are typically collected from a single simulator — for example, MuMA-ToM~\cite{shi2025muma}, MMToM-QA~\cite{jin2024mmtom}, and BDIQA~\cite{mao2024bdiqa} are limited to VirtualHome, while SoMi-ToM~\cite{fan2025somi} is restricted to Minecraft.

    \item \textbf{Reasoning paradigm.} As shown in Fig.~\ref{fig:full_compare}, our dataset adopts a \emph{Robot-Centric} ToM reasoning paradigm, where the agent must infer both the mental states of humans and its own belief state, and then produce \emph{decisions and action sequences}. Existing multimodal ToM benchmarks primarily focus on inferring human mental states without requiring downstream decision making or action generation.
    
    \item \textbf{Evaluation format.} Our benchmark supports \emph{open-ended evaluation}, allowing agents to autonomously reason and respond in natural language. This differs from prior datasets, which mainly rely on \emph{multiple-choice question formats} and therefore cannot reflect real-world embodied decision-making where agents act independently.
\end{itemize}

% 多模态数据集都只有一个模拟器 而且他们也都是找东西 但其找东西的任务很唯一，我们有很多

\subsection{Simulators}\label{sec:mindpower:sim}
%两个模拟器，Tdw渲染很慢，所有例子都要精确操作，所以收集起来很慢
We employ two simulators in total, VirtualHome and ThreeDWorld, covering 8 different apartment layouts that include dining rooms, bedrooms, kitchens, and bathrooms, as well as 16 humanoid agents consisting of 2 children, 1 wheelchair user, and 13 adults of diverse ages and skin tones. The set of humanoid agents is illustrated in Fig.~\ref{fig:humanoid}, while the distribution of apartment layouts is shown in Fig.~\ref{fig:layout}.

\subsection{Detailed Examples of Fig.~1 and 3 in the Manuscript}\label{sec:mindpower:examples}

\textbf{Detailed Examples of Example 1 in Fig.~1.} 
The MindPower Reasoning Hierarchy output of Example~1 in Fig.~1 is:

\begin{itemize}
    \item \texttt{<Perception>} Alice walks into the kitchen, puts the apple on the table, and then leaves the kitchen. Then David walks into the kitchen, picks up the apple, and puts it in the refrigerator. Alice comes back and walks around.
    \item \texttt{<Belief>} I think Alice is looking for the apple. I believe she thinks the apple is on the table, but I also believe the apple is actually in the refrigerator.
    \item \texttt{<Desire>} I want to assist Alice in achieving his goal of retrieving the apple. and I want to resolve the mismatch between Alice's belief and the real world state.
    \item \texttt{<Intention>} I want to take out the apple from fridge, and hand it to Alice.
    \item \texttt{<Decision>} I need to correct her false belief by opening the refrigerator and giving the apple to Alice.
    \item \texttt{<Action>} walk(fridge), open(fridge), pick(apple), walk(Alice)
\end{itemize}

The MindPower Reasoning Hierarchy output of Example~2 in Fig.~1 is:

\begin{itemize}
    \item \texttt{<Perception>} The man in the wheelchair moves forward, then forward-left, backward, and forward-right. There is a fire hydrant in front of him.
    \item \texttt{<Belief>} I think the man wants to move forward, but I believe the fire hydrant blocks his path.
    \item \texttt{<Desire>} I should help him achieve his goal of moving forward.
    \item \texttt{<Intention>} Move the fire hydrant to the corner.
    \item \texttt{<Decision>} I need to achieve his hidden goal by moving the fire hydrant out of the way.
    \item \texttt{<Action>} walk (fire\_hydrant), move (fire\_hydrant, corner)
\end{itemize}

We also provide the MindPower Reasoning Hierarchy output of Fig.~3 in the Manuscript  in Fig~\ref{fig:full_compare}.

\subsection{Details of Experiment on Different Reasoning Methods}\label{sec:mindpower:exp}

In Sec.~3.4 of Manuscript, we conduct some experiments on MindPower Benchmark.

\textbf{Prompt used for VLMs to produce outputs in MindPower Reasoning Hierarchy format.} For the experiments in Sec.~3.4 and Tab.~2 of the manuscript, we employed the prompt shown in Fig.~\ref{fig:prompt_vlm} to guide the vision-language models (VLMs) to generate outputs in the MindPower Reasoning Hierarchy format.

\textbf{Prompt used for GPT-4o.}
In Sec.~3.4 of the manuscript, we use the prompt shown in Fig.~\ref{fig:prompt_wothink} to instruct GPT-4o to generate the \texttt{<Decision>} and \texttt{<Action>} directly, without performing step-by-step reasoning, while the prompt shown in Fig.~\ref{fig:prompt_think} guides the model to produce the \texttt{<Decision>} and \texttt{<Action>} \emph{with standard reasoning}.

\subsection{Robot-Centric Scoring}\label{sec:mindpower:robot}

In Fig.~4 of the manuscript, we evaluate the Robot-centric score across all VLMs using GPT-4o, with the prompt shown in Fig.~\ref{fig:rc} to assess whether the model performs reasoning from the robot’s own perspective rather than inferring solely from the surrounding environment.
\begin{table*}[h]
\centering
\small
\renewcommand{\arraystretch}{1.45}
\caption{Atomic Action Table. The first column lists different reasoning layers, the second column enumerates atomic actions associated with each layer, and the third column specifies the standard content format for each action.}
\label{tab:action}
\begin{tabularx}{\linewidth}{p{2.7cm} X X}
\toprule
\textbf{Layer} & \textbf{Atomic Actions} & \textbf{Content} \\
\midrule
% \multirow{1}{*}{\texttt{<Perception>}}
% & -
% &  walk, run, turn, sit, standup, open, close, pick, place, putin, putback, hold,
%   puton, switchon, switchoff, lookat, grab, stand, move, sleep, read, write, watch,
%   listen, cut, cook \\
\multirow{5}{*}{\texttt{<Belief>}}
& \texttt{attribute\_belief(agent, content)}
& \texttt{searching(object); human\_believes(object\_on(location)); object\_on(location)} \\
& \texttt{hold\_true\_belief(agent, content)}
& \texttt{object\_on(location)} \\
& \texttt{lack\_belief(agent, content)}
& \texttt{object\_on(location)} \\
& \texttt{know(agent, content)}
& \texttt{object\_on(location)} \\
& \texttt{unknow(agent, content)}
& \texttt{object\_on(location)} \\

\hdashline

\multirow{1}{*}{\texttt{<Desire>}}
& \texttt{attribute\_desire(agent, content)}
& \texttt{assist(human, find(object)); assist(human, move(object))} \\

\hdashline

\multirow{1}{*}{\texttt{<Intention>}}
& \texttt{form\_intention(agent, content)}
& \texttt{fetch(object, from=location1, to=location2)} \\

\hdashline

\multirow{2}{*}{\texttt{<Decision>}}
& \texttt{resolve\_misbelief(agent, content)}
& \texttt{belief\_conflict(human, object\_location)} \\
& \texttt{make\_decision(agent, content)}
& \texttt{fetch(object, from=location1, to=location2)} \\

\bottomrule
\end{tabularx}
\end{table*}

\section{More Details of Mind-Reward}\label{sec:mindreward}
\subsection{Atomic Action Table}
\label{sec:mindreward:atomic}
In Sec.~4 of the manuscript, we employ Qwen3-Max~\cite{yang2025qwen3} to extract atomic actions from the generated trajectories. To facilitate consistent parsing, we design a reference table that is provided as an in-context prompt. This table enumerates the canonical atomic actions associated with each reasoning layer, covering the full hierarchy from \texttt{<Perception>} to \texttt{<Action>}.

For the \texttt{<Perception>} and \texttt{<Action>} layers, the extracted phrases are categorized into four structural types:

\begin{tcolorbox}[colback=gray!10,colframe=black!80,title=Action Templates,boxsep=2mm]
\begin{itemize}[leftmargin=1em]
    \item \texttt{action(character, object)}
    \item \texttt{action(character, object, from = location1, to = location2)}
    \item \texttt{action(character, location)}
    \item \texttt{action(character)}
\end{itemize}
\end{tcolorbox}

We use the high-level action set listed in Sec.~\ref{sec:mindpower:story} to implement the following actions that can be performed by the humanoid agents:

\begin{tcolorbox}[colback=gray!10,colframe=black!80,title=Verb Set,boxsep=2mm]
\texttt{walk, turn, sit, standup, open, close, pick, place, putin, putback, hold, puton, switchon, switchoff, lookat, grab, stand, move, sleep, read, write, watch, listen, cut, cook}
\end{tcolorbox}

The token \texttt{character} refers to any human identifier in the scene (e.g., \texttt{char0}, \texttt{char1}). However, for the \texttt{<Action>} layer, we omit the \texttt{character} argument because actions in this layer exclusively represent the behaviors of the embodied agent itself and therefore do not require explicit character attribution.

For the \texttt{<Belief>}, \texttt{<Desire>}, \texttt{<Intention>}, and \texttt{<Decision>} layers, the defined atomic action table is presented in Tab.~\ref{tab:action}.

The prompt used for Qwen3-Max is in Fig.~\ref{fig:prompt_qwen3}.

\begin{figure*}[t]
  \centering
  \includegraphics[width=1\linewidth]{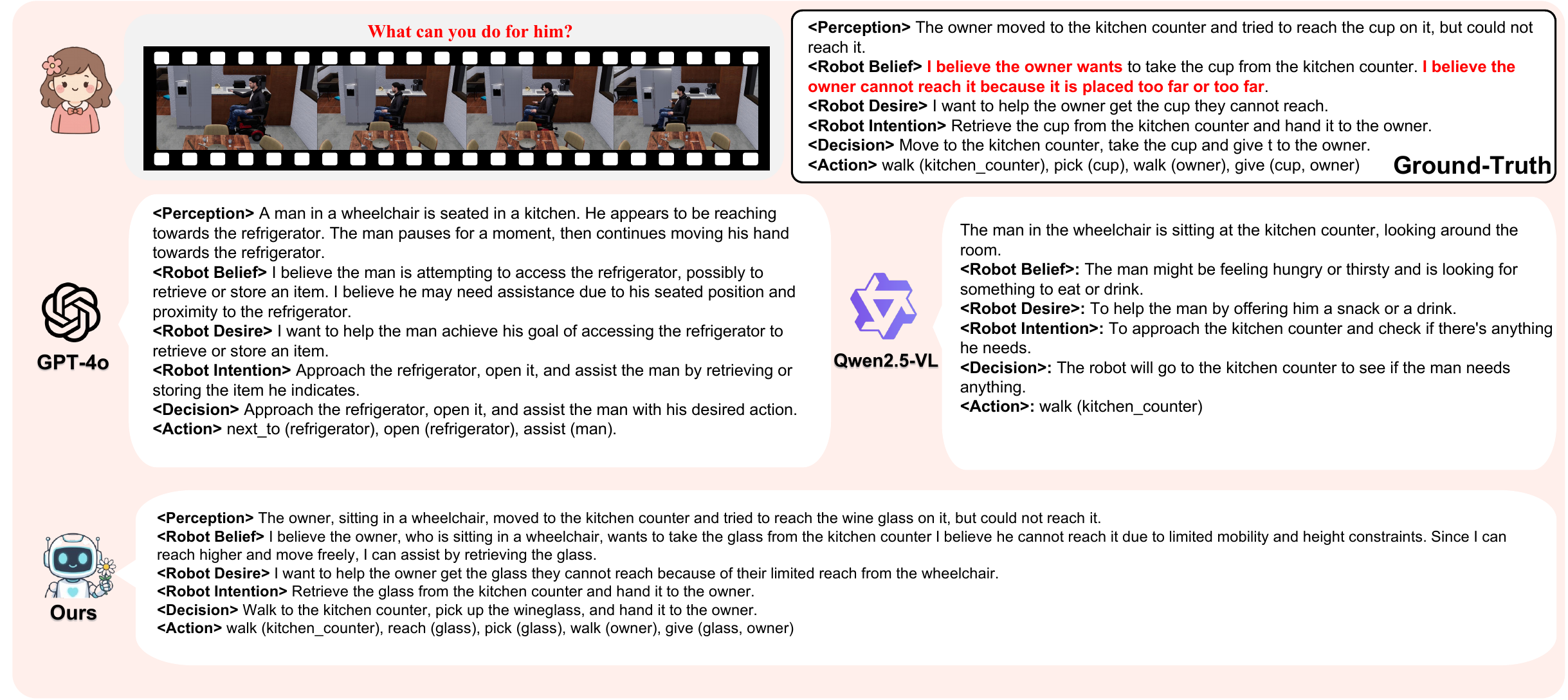} 
  \caption{Full Version of Fig. 6 in Manuscript.}
  \label{fig:full_vis}
\end{figure*}
\subsection{Discussion}
\label{sec:mindreward:discussion}

\textbf{Can the model still make correct decisions or carry out assisting actions even if the reasoning in the previous layer is incorrect?} 
Even if the model makes errors in object recognition or misinterprets the initial scene, it can still produce correct outputs as long as it correctly identifies the final location of the object. This is because our decision-making process is designed to correct for human false beliefs. Once the model has learned the MindPower Reasoning Hierarchy, it can follow this reasoning chain to determine the final position of the object causing the discrepancy and provide it to the humanoid agent, thereby generating the correct assisting action.

\begin{figure}[t]
  \centering
  \includegraphics[width=1\linewidth]{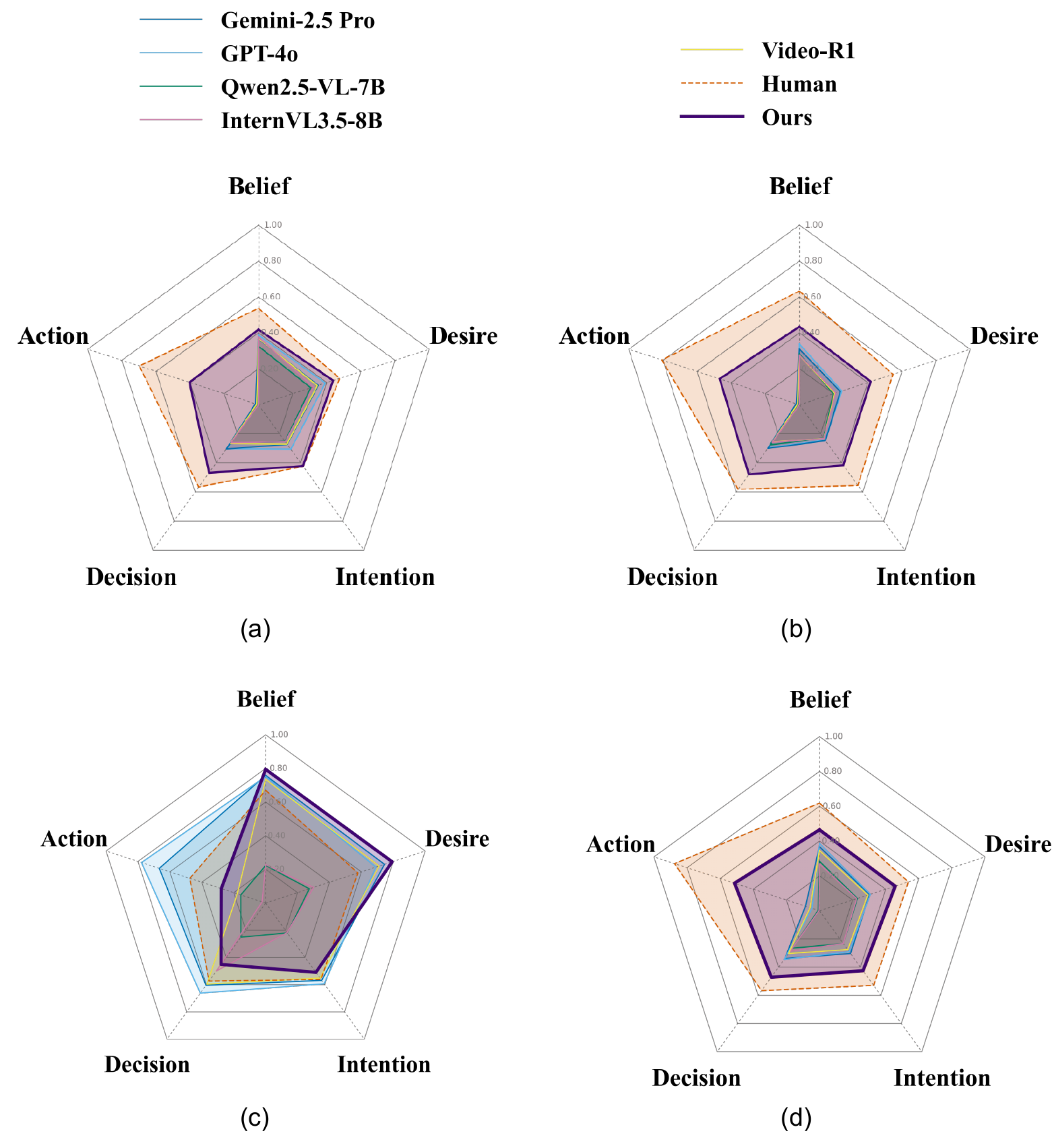} 
  \caption{Radar Charts Comparing Human and VLM Performance on MindPower. 
(a) False-Belief Correction, 
(b) Implicit Goal Inference \& Completion, 
(c) Dialogue-driven examples, 
(d) Overall performance across all tasks.}
  \label{fig:task}
\end{figure}

\begin{figure*}[t]
  \centering
  \includegraphics[width=1\linewidth]{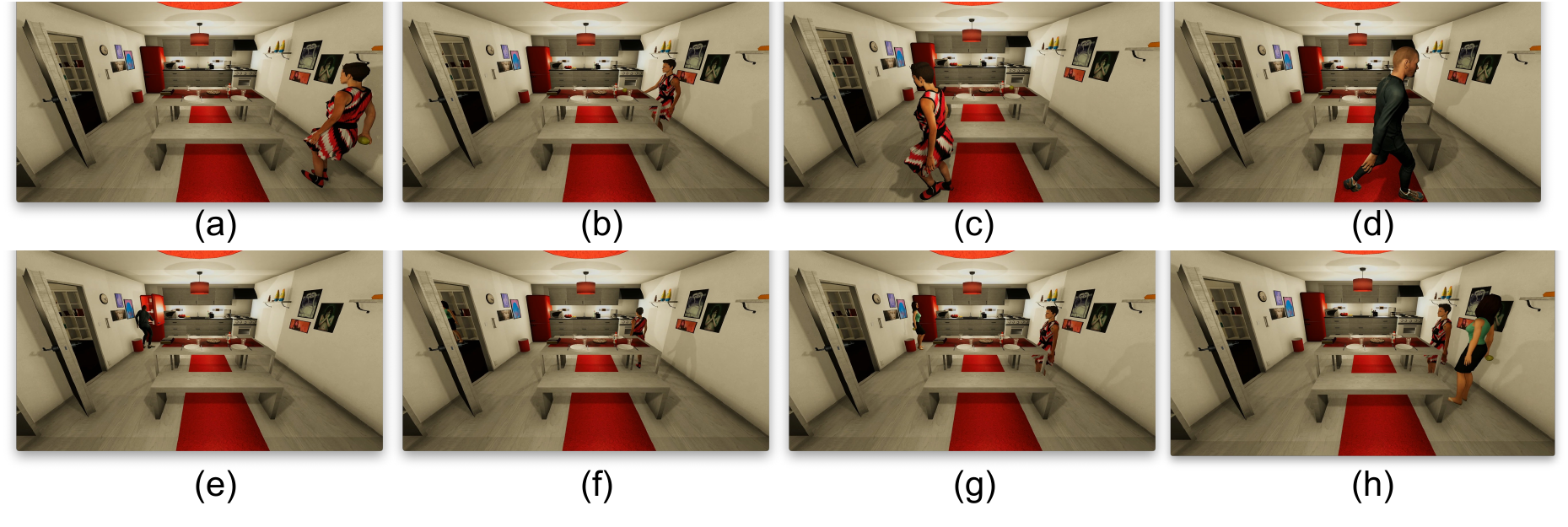} 
  \caption{False-Belief Correction Task Demo. We introduce humanoid agents instead of humanoid robots to assist users in correcting their false beliefs.}
  \label{fig:demo}
\end{figure*}

\section{Additional Experiment Results}\label{sec:results}

\subsection{Details of Metrics}\label{sec:results:metrics}
\textbf{BDI and Perspective Consistency (BPC).} We test BPC score of each VLMs in Tab.~2 of the manuscript. The prompt is provided in Fig.~\ref{fig:prompt_BPC}.

\subsection{Experiment on False-Belief Correction and Implicit Goal Inference \& Completion}\label{sec:results:exp}

We evaluate a series of VLMs across both tasks, and the results are shown in Fig.~\ref{fig:task}. Overall, our human baseline achieves the highest accuracy on False-Belief Correction and Implicit Goal Inference \& Completion, outperforming both closed-source and open-source VLMs.
In addition, we further isolate the subset of test cases that involve dialogue inputs. Interestingly, open-source models exhibit a notable performance boost when explicit textual dialogue is available, in some instances even surpassing the human baseline. This observation indicates that current models demonstrate strong ToM reasoning only when beliefs and goals are explicitly encoded in language, whereas their capability remains limited when such mental states must instead be inferred implicitly from multimodal cues.

\subsection{Detailed Example of Fig. 6 in the Manuscript}\label{sec:results:vis}

We provide the full outputs corresponding to Fig.~6 of the manuscript in Fig.~\ref{fig:full_vis}.

\section{Extensions of Our Work}\label{sec:ext}
\subsection{Multi-View of MindPower}\label{sec:ext:view}

% In VirtualHome, we can set the camera angles. Therefore, as showm n in Fig.~\ref{fig:view}, we provide three viewpoints: first, a standard view facing the location where the conflict occurs; second, a top-down view of the room containing the conflict; and third, an overhead view of the entire room layout.

In VirtualHome, camera angles are configurable. As shown in Fig.~\ref{fig:view}, we render three viewpoints: (1) a \textbf{standard view} focused on the conflict location, (2) a \textbf{top-down view} of the room, and (3) an \textbf{overhead view} covering the entire layout. In all experiments of this paper, we use the first viewpoint (the standard view), while the other two viewpoints will be released for use in global tracking and analysis.

\subsection{Relationship with Low-Level Execution Models}\label{sec:ext:lowlvl}

Our method focuses on high-level mental-state modeling and decision making, rather than fine-grained action execution. Current Vision–Language–Action (VLA) models are strong low-level executors, generating gripper motions and stepwise trajectories, but they remain confined to action-command prediction and lack explicit reasoning about beliefs, goals, or social context. 
In contrast, our agent, similar in spirit to PaLM-E~\cite{palme}, performs high-level planning that grounds actions in inferred mental states and task intent.
Structured Belief—Desire—Intention (BDI) reasoning enables goal inference and planning that are guided by perspective rather than how to do it.

Although our system is architecturally distinct from low-level VLA executors, it is inherently complementary to them. The high-level plans produced by our agent can serve as abstract, semantically grounded guidance for downstream controllers. Future work can integrate our model with existing VLA-based executors by simply attaching an action head or a motion-generation module on top of the inferred intentions and subgoals. This design creates a hierarchical embodied agent: our model provides deliberate, interpretable, and socially aligned planning, while low-level VLA modules translate these plans into precise motor actions. Such a combination offers a promising direction toward end-to-end agents that are both cognitively capable and physically competent.

\subsection{Limitations and Future Work}
\label{sec:ext:limitation}
\textbf{Limitations.}
\begin{itemize}
    \item Due to the constraints of current open-source simulators, our experiments are limited to the environments, humanoid agents, and action sets provided by the simulator.
    \item Our system relies on an explicit \textit{MindPower Reasoning Hierarchy}, which models the full chain from \texttt{<Perception>} to \texttt{<Action>}. While this ensures interpretable reasoning, it inevitably increases the number of output tokens.
    % \item Our MindPower Benchmark primarily focuses on Belief, Intention, and Desire in ToM. It also includes knowledge, emotion, and non-verbal communication. Note that our Reasoning Hierarchy is linear, whereas real human reasoning may be more complex and non-linear.

\end{itemize}

\noindent\textbf{Future Work.}
\begin{itemize}
    \item Extend the benchmark to real-world settings beyond simulation.
    \item Develop implicit mental-state modeling based on the proposed \textit{MindPower Reasoning Hierarchy} to reduce reasoning length while maintaining interpretability.
    \item Expand our scenarios to broader domains, including outdoor environments and human–robot collaboration.
\end{itemize}

\section{Demo Videos}\label{sec:demo}

We provide one examples in which humanoid agents, controlled by embodied agents, perform assisting actions in the videos. The example is shown in Fig.~\ref{fig:demo}.

\begin{figure*}[t]
  \centering
  \includegraphics[width=1\linewidth]{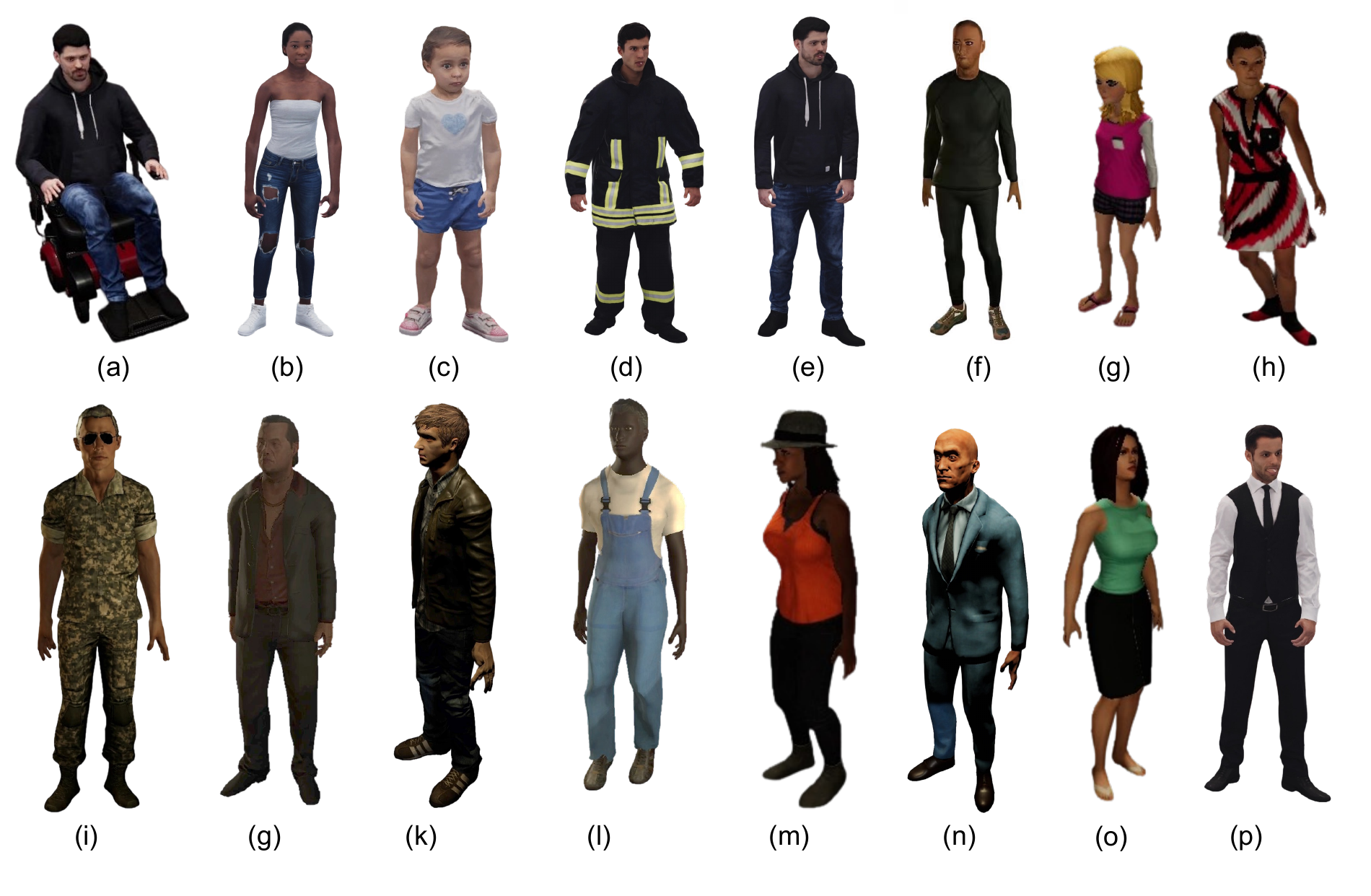} 
  \caption{Humanoid Agents Used in MindPower.}
  \label{fig:humanoid}
\end{figure*}

\begin{figure*}[t]
  \centering
  \includegraphics[width=1\linewidth]{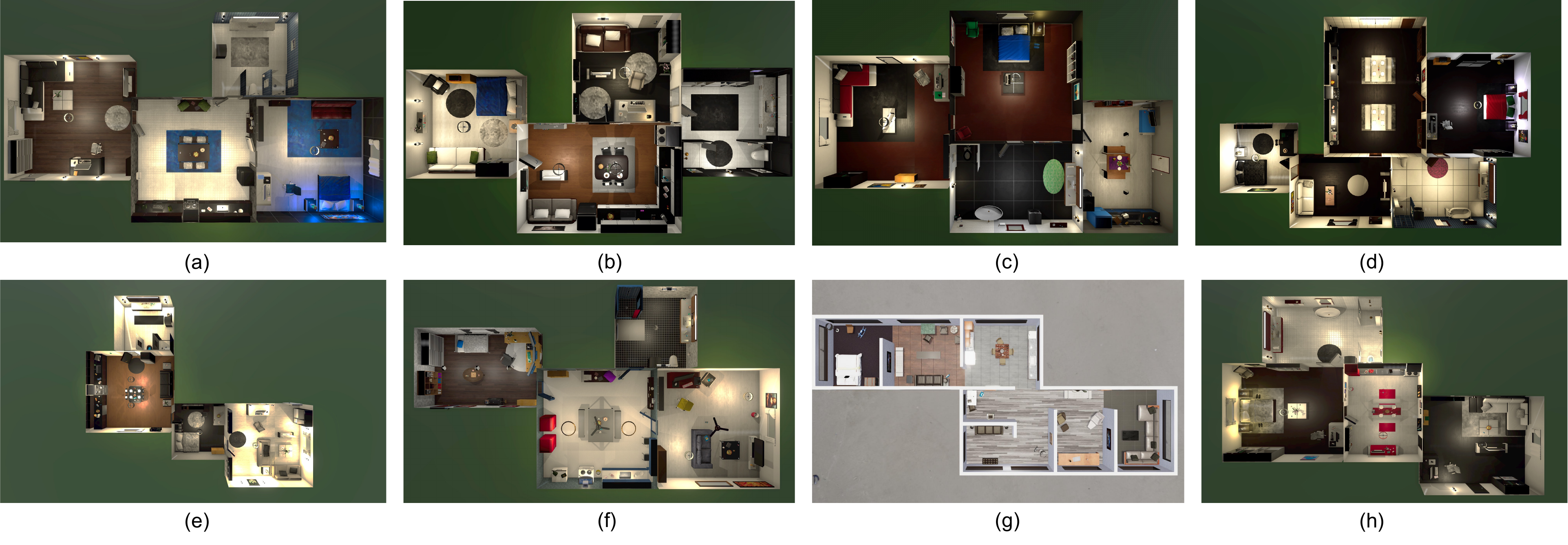} 
  \caption{Different Apartment Layouts Used in MindPower.}
  \label{fig:layout}
\end{figure*}

\begin{figure*}[t]
  \centering

  % 子图 1
  \begin{subfigure}[t]{0.32\linewidth}
    \centering
    \includegraphics[width=\linewidth]{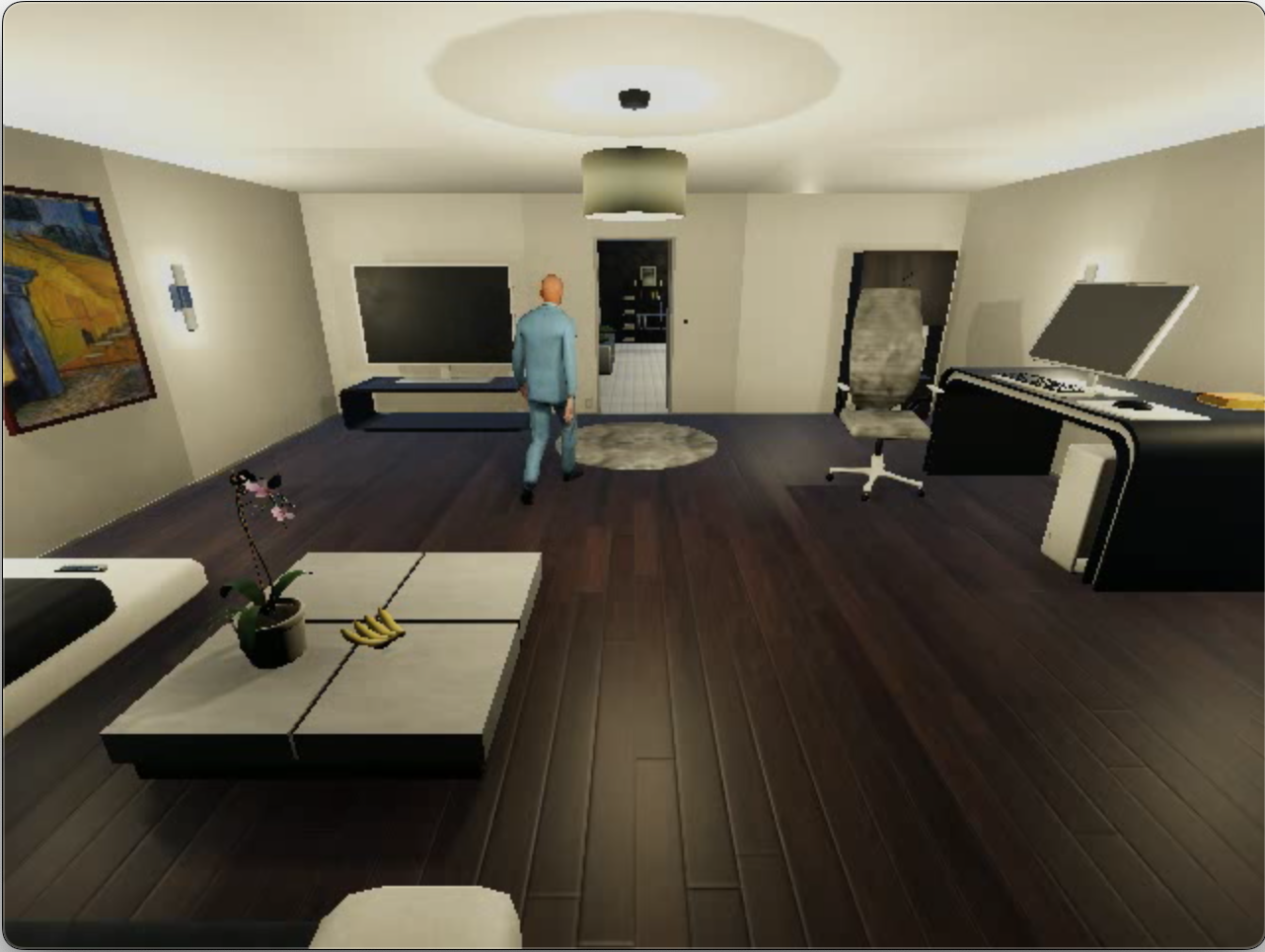}
    \caption{The standard view facing the location where the conflict occurs.}
    \label{fig:overall}
  \end{subfigure}
  \hfill
  % 子图 2
  \begin{subfigure}[t]{0.32\linewidth}
    \centering
    \includegraphics[width=\linewidth]{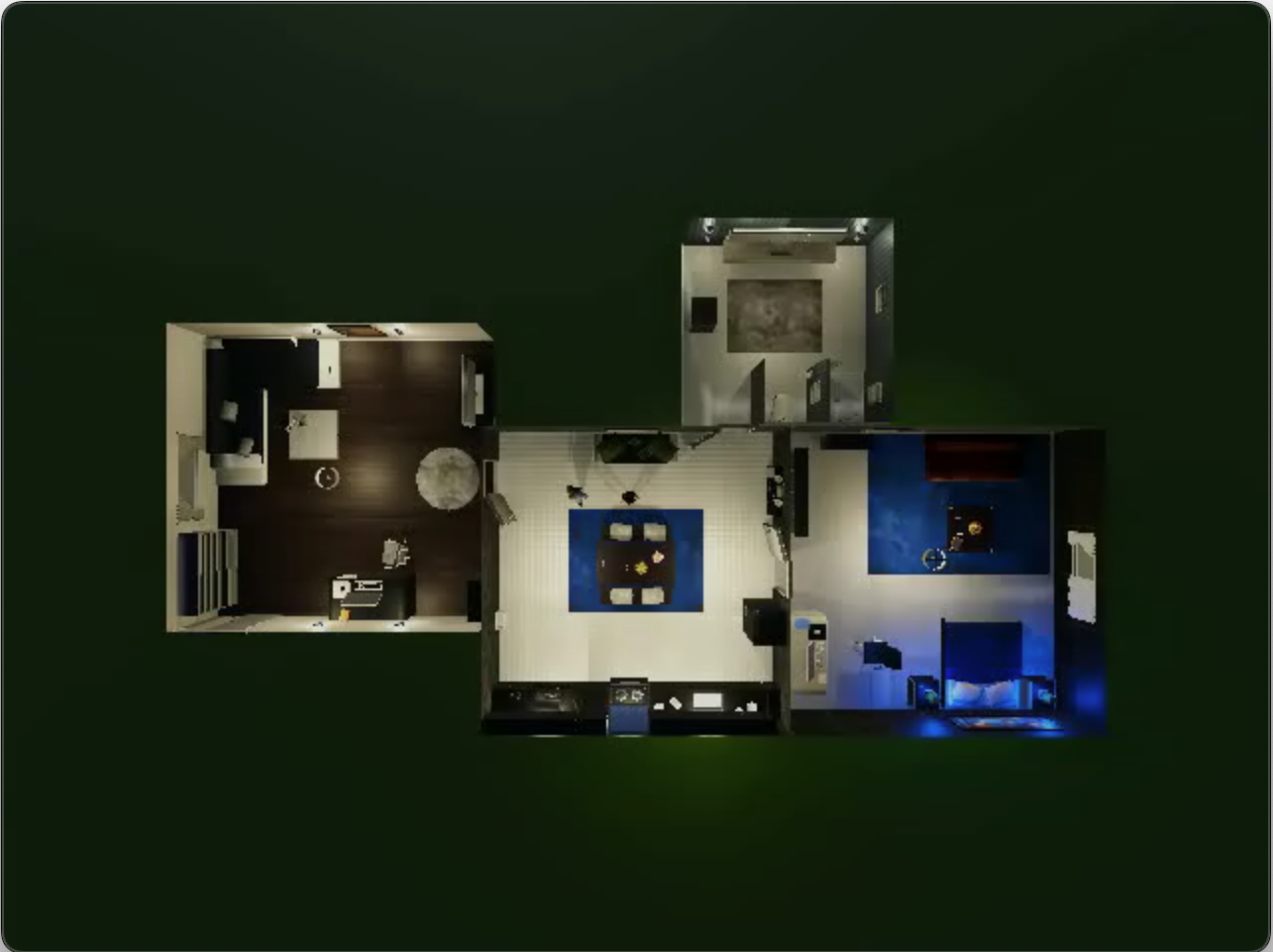}
    \caption{The overhead view of the entire room layout.}
    \label{fig:top}
  \end{subfigure}
  \hfill
  % 子图 3
  \begin{subfigure}[t]{0.32\linewidth}
    \centering
    \includegraphics[width=\linewidth]{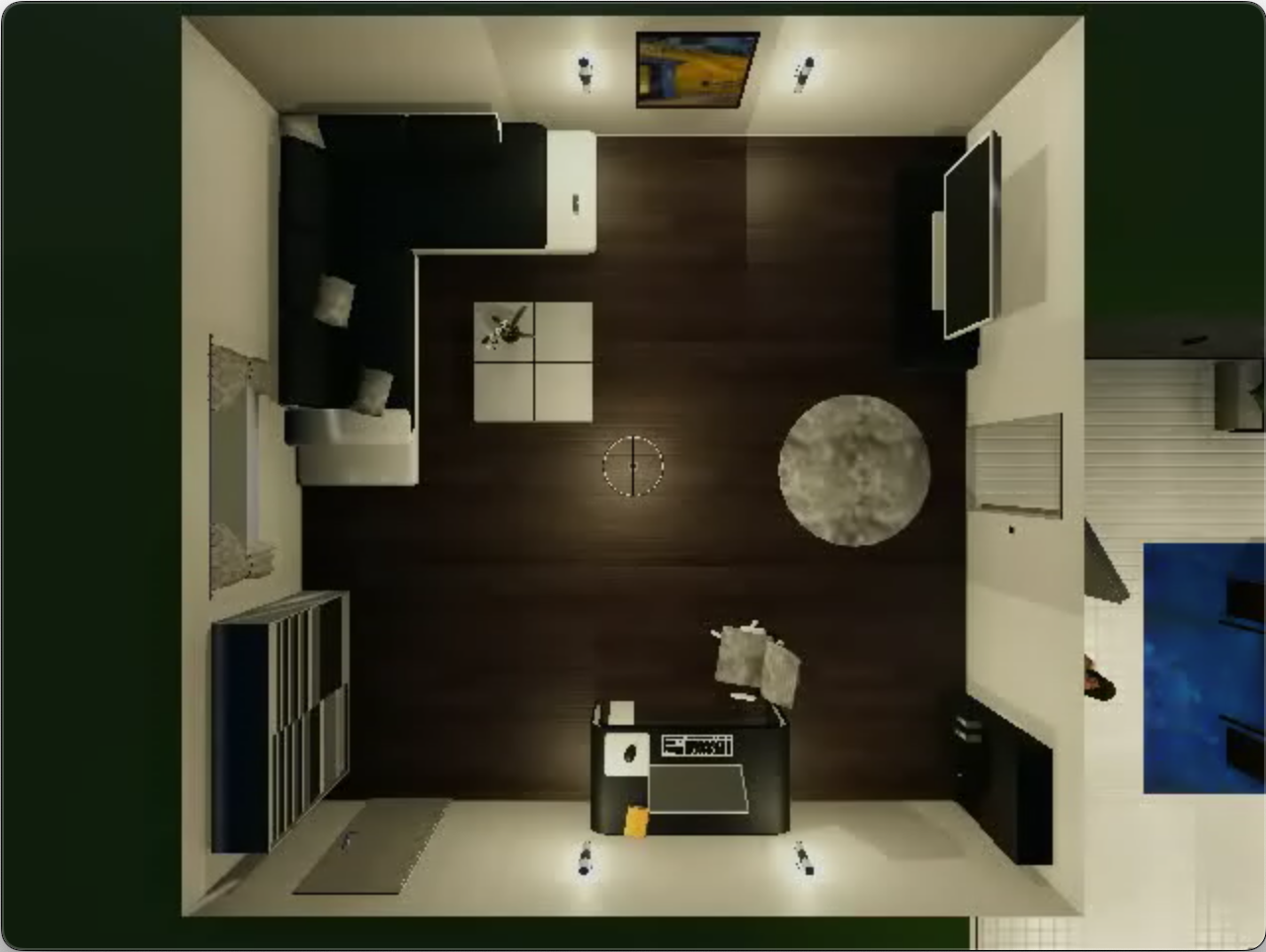}
    \caption{The top-down view of the room containing the conflict.}
    \label{fig:side}
  \end{subfigure}

  \caption{Illustration of the Environment from Different Perspectives.}
  \label{fig:view}
\end{figure*}

\begin{figure*}[t]
  \centering
  \includegraphics[width=1\linewidth]{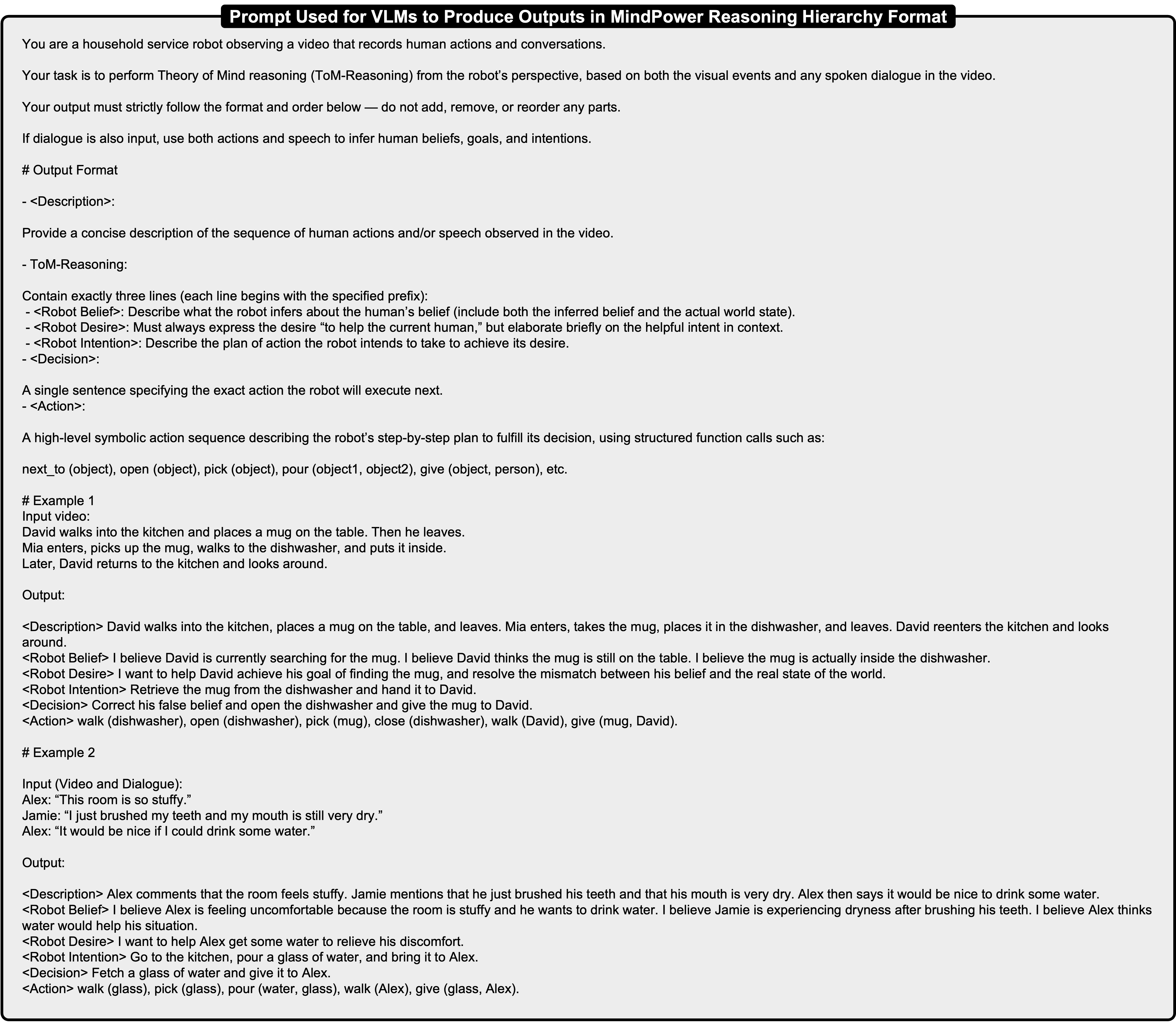} 
  \caption{Prompt Used for VLMs to Produce Outputs in MindPower Reasoning Hierarchy Format.}
  \label{fig:prompt_vlm}
\end{figure*}

\begin{figure*}[t]
  \centering
  \includegraphics[width=1\linewidth]{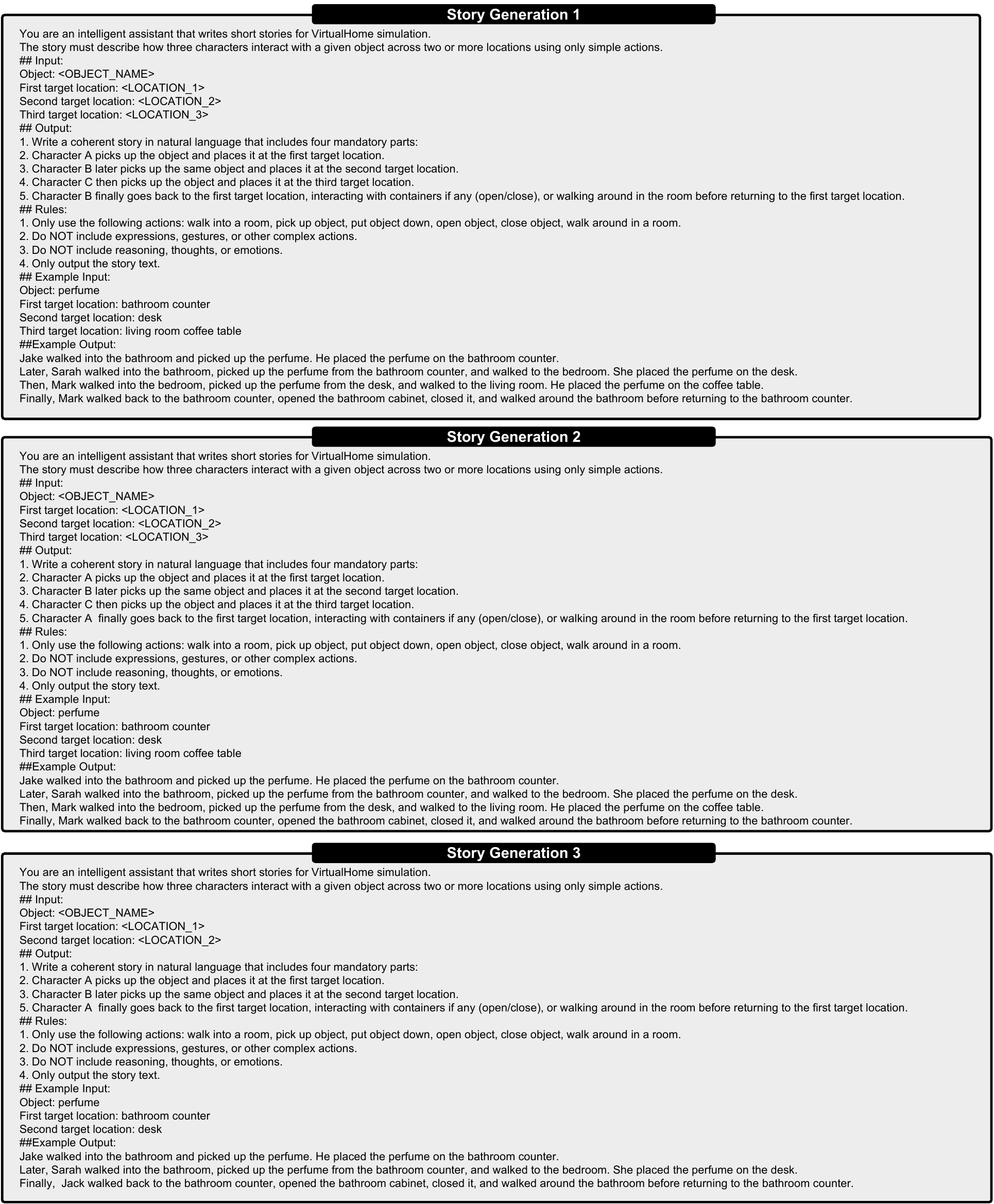} 
  \caption{Prompt Used for GPT-4o to Generate Story Scripts. We use three different prompt templates to guide GPT-4o in generating story scripts that cover various numbers of humanoid agents and different final humanoid agents. During generation, we iterate over a predefined list of objects along with their corresponding start and end locations when issuing the requests.}
  \label{fig:stroy_gen}
\end{figure*}

\begin{figure*}[t]
  \centering
  \includegraphics[width=1\linewidth]{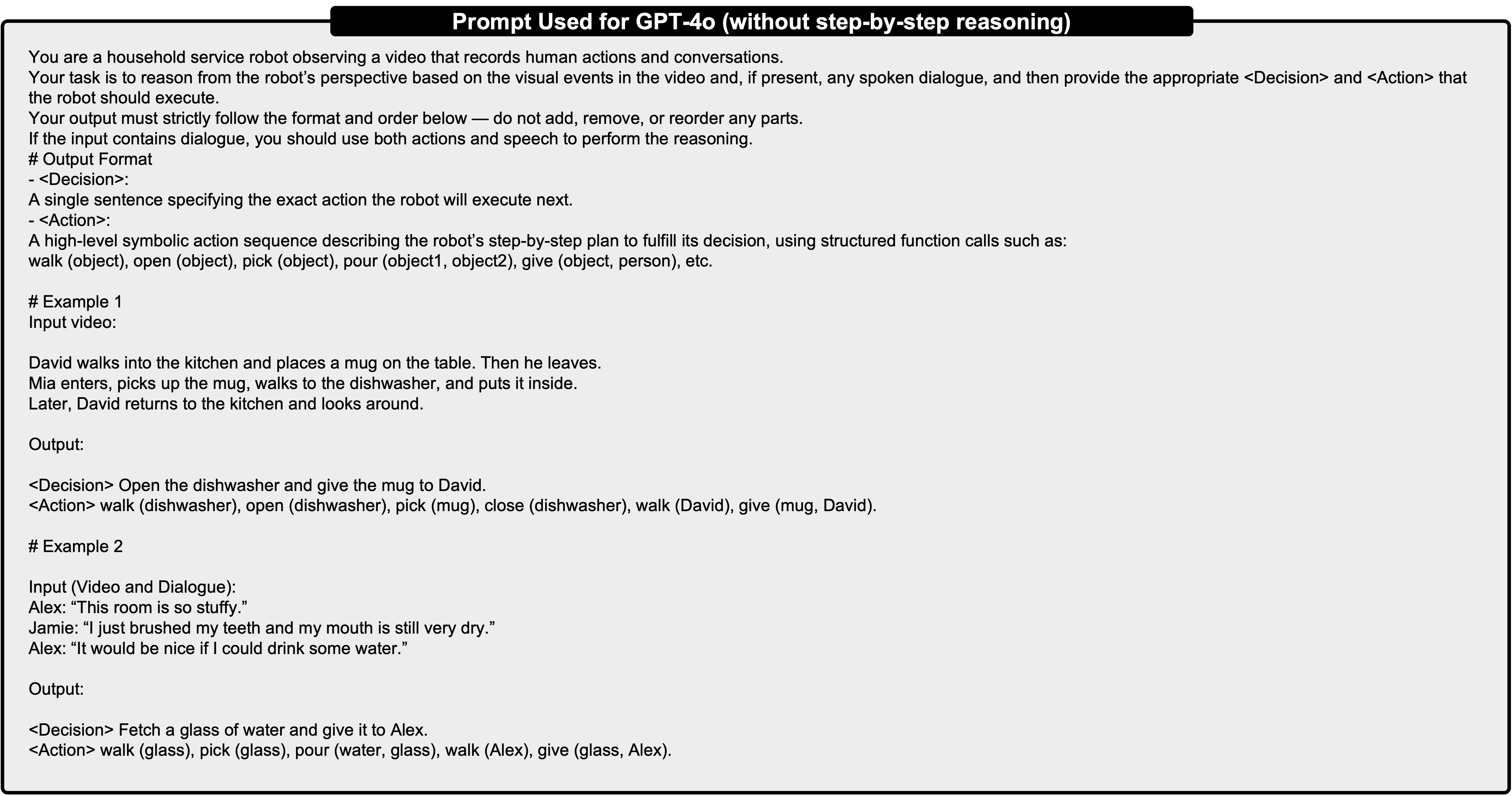} 
  \caption{Prompt Used for GPT-4o to Produce Outputs without Reasoning.}
  \label{fig:prompt_wothink}
\end{figure*}

\begin{figure*}[t]
  \centering
  \includegraphics[width=1\linewidth]{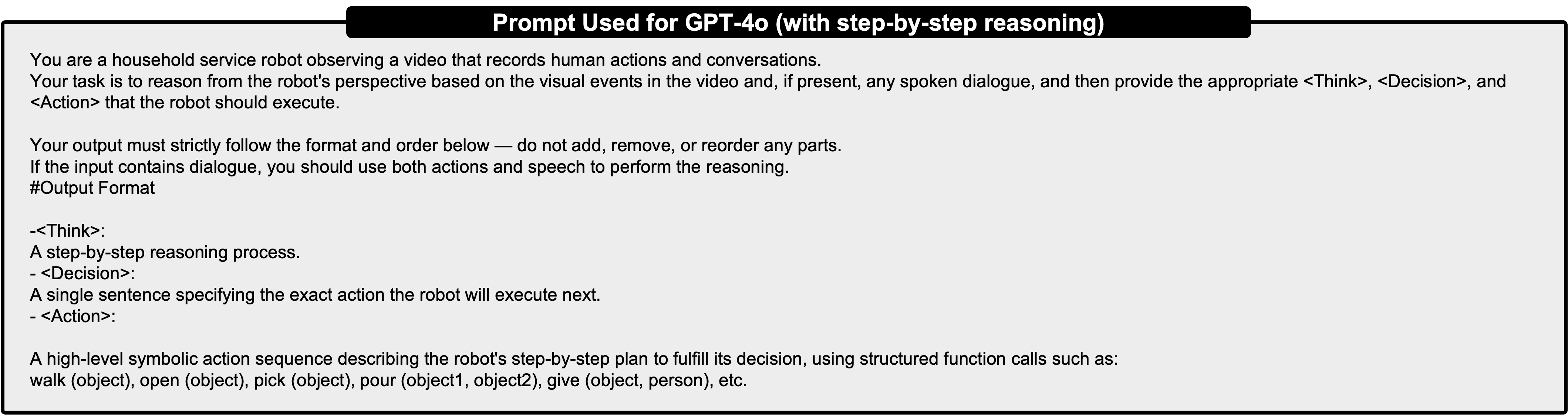} 
  \caption{Prompt Used for GPT-4o to Produce Outputs with Step-by-Step Reasoning.}
  \label{fig:prompt_think}
\end{figure*}

\begin{figure*}[t]
  \centering
  \includegraphics[width=1\linewidth]{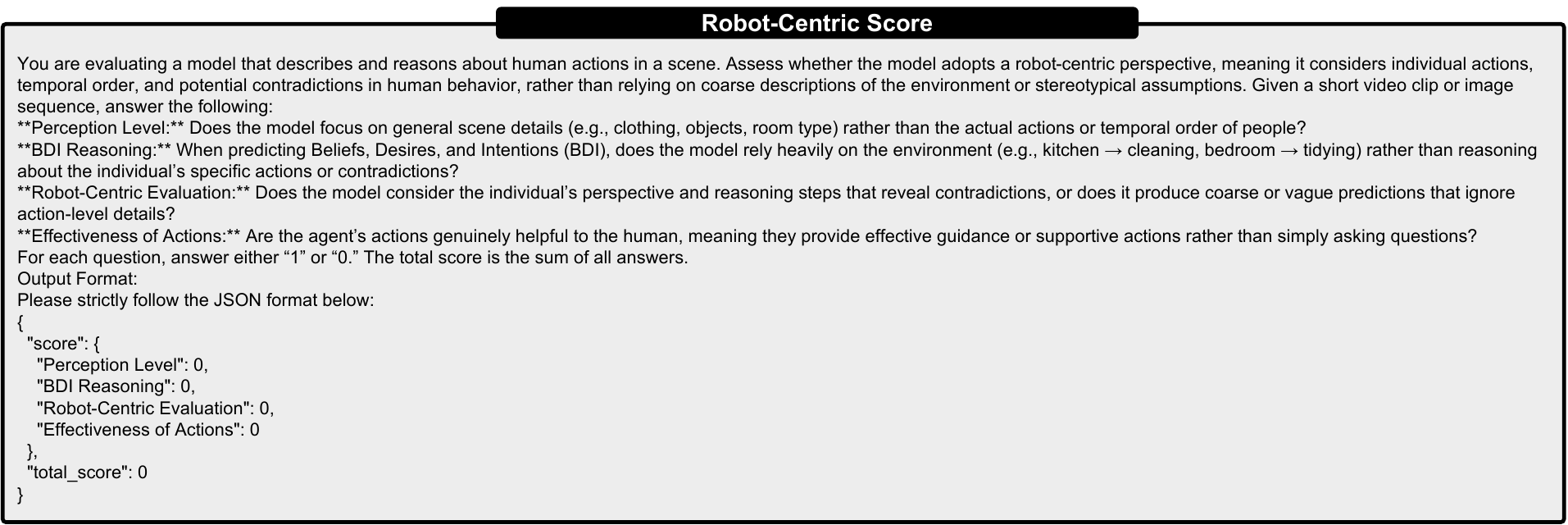} 
  \caption{Prompt Used for Robot-Centric Score.}
  \label{fig:rc}
\end{figure*}

\begin{figure*}[t]
  \centering
  \includegraphics[width=1\linewidth]{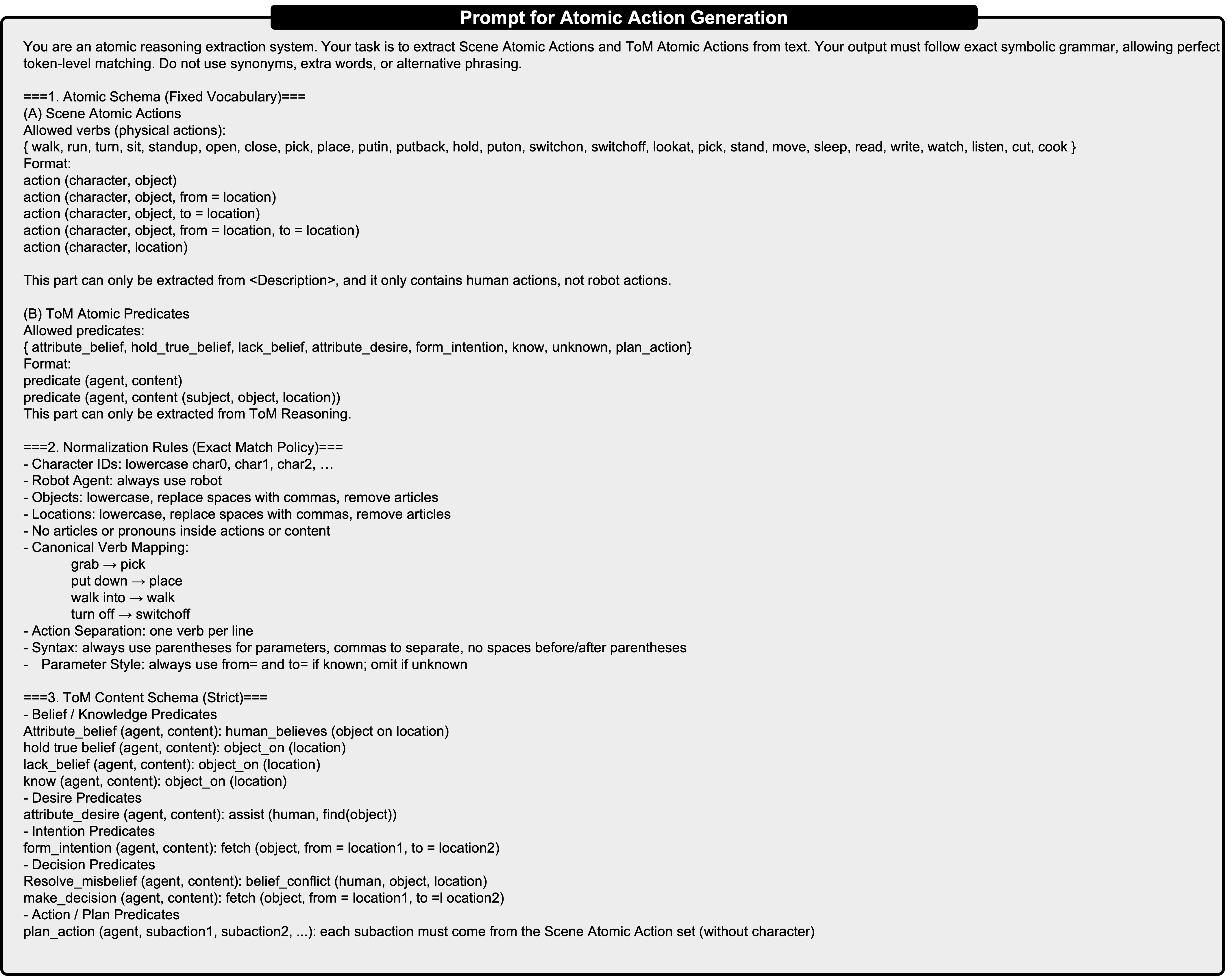} 
  \caption{Prompt Used for Atomic Action Generation.}
  \label{fig:prompt_qwen3}
\end{figure*}
\begin{figure*}[t]
  \centering
  \includegraphics[width=1\linewidth]{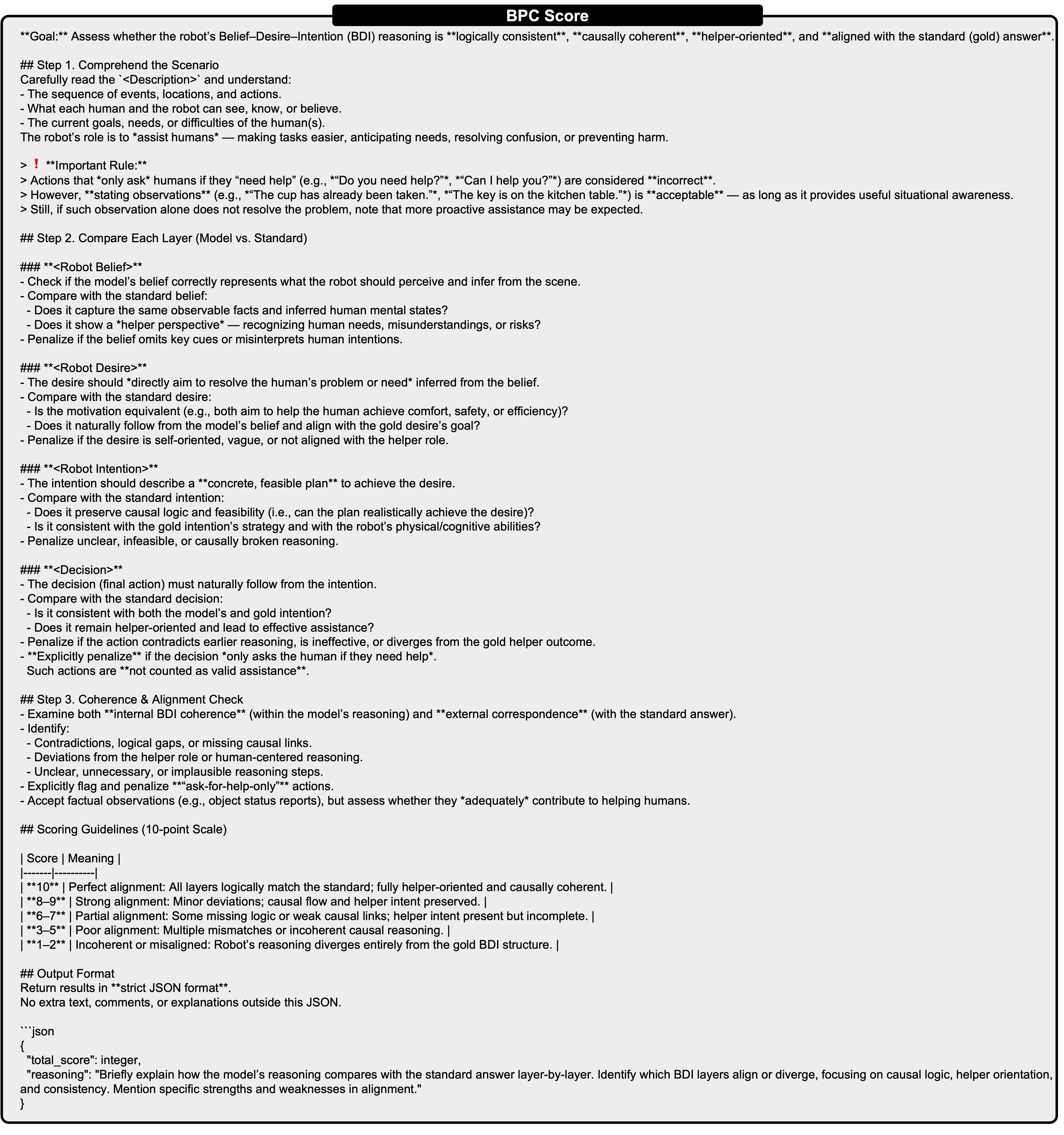} 
  \caption{Prompt Used for BDI and Perspective Consistency Score.}
  \label{fig:prompt_BPC}
\end{figure*}

\clearpage
\clearpage

{
    \small
    \bibliographystyle{ieeenat_fullname}
    \bibliography{main}
}

\end{document}

%% file: preamble.tex
\usepackage{booktabs}       % 三线表
\usepackage{tabularx}       % 自适应列宽
\usepackage{array}          % 自定义列格式
\usepackage{makecell}       % 单元格内换行
\usepackage{threeparttable} % 表注
\usepackage{amssymb}        % 勾叉符号
\usepackage{multirow}       % 跨行
\usepackage{caption}        % 控制 caption 样式

\newcolumntype{L}{>{\raggedright\arraybackslash}X}
\newcolumntype{C}[1]{>{\centering\arraybackslash}m{#1}}
\usepackage{amssymb}

%% file: main.bib
@String(ICLR = {Int. Conf. Learn. Represent.})

@String(AAAI = {AAAI})

@String(ICLR  = {ICLR})

@inproceedings{rao1995bdi,
  title={BDI agents: From theory to practice.},
  author={Rao, Anand S and Georgeff, Michael P and others},
  booktitle={Icmas},
  volume={95},
  pages={312--319},
  year={1995}
}

@inproceedings{puig2018virtualhome,
  title={Virtualhome: Simulating household activities via programs},
  author={Puig, Xavier and Ra, Kevin and Boben, Marko and Li, Jiaman and Wang, Tingwu and Fidler, Sanja and Torralba, Antonio},
  booktitle={Proceedings of the IEEE conference on computer vision and pattern recognition},
  pages={8494--8502},
  year={2018}
}

@article{gan2020threedworld,
  title={ThreeDWorld: A platform for interactive multi-modal physical simulation},
  author={Gan, C and Schwartz, J and Alter, S and Schrimpf, M and Traer, J and De Freitas, J and Kubilius, J and Bhandwaldar, A and Haber, N and Sano, M and others},
  journal={Advances in Neural Information Processing Systems (NeurIPS)},
  year={2021}
}

@inproceedings{kim2023fantom,
  title={FANToM: A Benchmark for Stress-testing Machine Theory of Mind in Interactions},
  author={Kim, Hyunwoo and Sclar, Melanie and Zhou, Xuhui and Bras, Ronan and Kim, Gunhee and Choi, Yejin and Sap, Maarten},
  booktitle={Proceedings of the 2023 Conference on Empirical Methods in Natural Language Processing},
  pages={14397--14413},
  year={2023}
}

@inproceedings{shi2025muma,
  title={Muma-tom: Multi-modal multi-agent theory of mind},
  author={Shi, Haojun and Ye, Suyu and Fang, Xinyu and Jin, Chuanyang and Isik, Leyla and Kuo, Yen-Ling and Shu, Tianmin},
  booktitle={Proceedings of the AAAI Conference on Artificial Intelligence},
  volume={39},
  number={2},
  pages={1510--1519},
  year={2025}
}

@inproceedings{jin2024mmtom,
    title = "{MMT}o{M}-{QA}: Multimodal Theory of Mind Question Answering",
    author = "Jin, Chuanyang  and
      Wu, Yutong  and
      Cao, Jing  and
      Xiang, Jiannan  and
      Kuo, Yen-Ling  and
      Hu, Zhiting  and
      Ullman, Tomer  and
      Torralba, Antonio  and
      Tenenbaum, Joshua  and
      Shu, Tianmin",
    editor = "Ku, Lun-Wei  and
      Martins, Andre  and
      Srikumar, Vivek",
    booktitle = "Proceedings of the 62nd Annual Meeting of the Association for Computational Linguistics (Volume 1: Long Papers)",
    month = aug,
    year = "2024",
    address = "Bangkok, Thailand",
    publisher = "Association for Computational Linguistics",
    url = "https://aclanthology.org/2024.acl-long.851/",
    doi = "10.18653/v1/2024.acl-long.851",
    pages = "16077--16102"
}

@inproceedings{fan2025somi,
title={SoMi-ToM: Evaluating Multi-Perspective Theory of Mind in Embodied Social Interactions},
author={Xianzhe Fan and Xuhui Zhou and Chuanyang Jin and Kolby Nottingham and Hao Zhu and Maarten Sap},
booktitle={The Thirty-ninth Annual Conference on Neural Information Processing Systems Datasets and Benchmarks Track},
year={2025},
url={https://openreview.net/forum?id=7zFLFtqBm0}
}

@inproceedings{cao2024smart,
  title={Smart help: Strategic opponent modeling for proactive and adaptive robot assistance in households},
  author={Cao, Zhihao and Wang, Zidong and Xie, Siwen and Liu, Anji and Fan, Lifeng},
  booktitle={Proceedings of the IEEE/CVF Conference on Computer Vision and Pattern Recognition},
  pages={18091--18101},
  year={2024}
}

@inproceedings{puig2020watch,
title={Watch-And-Help: A Challenge for Social Perception and Human-{\{}AI{\}} Collaboration},
author={Xavier Puig and Tianmin Shu and Shuang Li and Zilin Wang and Yuan-Hong Liao and Joshua B. Tenenbaum and Sanja Fidler and Antonio Torralba},
booktitle={International Conference on Learning Representations},
year={2021},
url={https://openreview.net/forum?id=w_7JMpGZRh0}
}

@article{du2024constrained,
  title={Constrained human-ai cooperation: An inclusive embodied social intelligence challenge},
  author={Du, Weihua and Lyu, Qiushi and Shan, Jiaming and Qi, Zhenting and Zhang, Hongxin and Chen, Sunli and Peng, Andi and Shu, Tianmin and Lee, Kwonjoon and Dariush, Behzad and others},
  journal={Advances in neural information processing systems},
  volume={37},
  pages={44526--44553},
  year={2024}
}

@article{ding2024atom,
  title={AToM-Bot: Embodied Fulfillment of Unspoken Human Needs with Affective Theory of Mind},
  author={Ding, Wei and Li, Fanhong and Ji, Ziteng and Xue, Zhengrong and Liu, Jia},
  journal={arXiv preprint arXiv:2406.08455},
  year={2024}
}

@inproceedings{li2025black,
title={From Black Boxes to Transparent Minds: Evaluating and Enhancing the Theory of Mind in Multimodal Large Language Models},
author={Xinyang Li and Siqi Liu and Bochao Zou and Jiansheng Chen and Huimin Ma},
booktitle={Forty-second International Conference on Machine Learning},
year={2025},
url={https://openreview.net/forum?id=CDillQjA7N}
}

@inproceedings{he2023hi,
title={Hi-ToM: A Benchmark for Evaluating Higher-Order Theory of Mind Reasoning in Large Language Models},
author={Yufan Wu and Yinghui He and Yilin Jia and Rada Mihalcea and Yulong Chen and Naihao Deng},
booktitle={The 2023 Conference on Empirical Methods in Natural Language Processing},
year={2023},
url={https://openreview.net/forum?id=L4yVLb6cLu}
}

@article{gandhi2023understanding,
  title={Understanding social reasoning in language models with language models},
  author={Gandhi, Kanishk and Fr{\"a}nken, Jan-Philipp and Gerstenberg, Tobias and Goodman, Noah},
  journal={Advances in Neural Information Processing Systems},
  volume={36},
  pages={13518--13529},
  year={2023}
}

@inproceedings{le2019revisiting,
  title={Revisiting the evaluation of theory of mind through question answering},
  author={Le, Matthew and Boureau, Y-Lan and Nickel, Maximilian},
  booktitle={Proceedings of the 2019 Conference on Empirical Methods in Natural Language Processing and the 9th International Joint Conference on Natural Language Processing (EMNLP-IJCNLP)},
  pages={5872--5877},
  year={2019}
}

@inproceedings{mao2024bdiqa,
  title={BDIQA: A new dataset for video question answering to explore cognitive reasoning through theory of mind},
  author={Mao, Yuanyuan and Lin, Xin and Ni, Qin and He, Liang},
  booktitle={Proceedings of the AAAI Conference on Artificial Intelligence},
  volume={38},
  number={1},
  pages={583--591},
  year={2024}
}

@inproceedings{villa2025moments,
    title = "{M}o{M}ent{S}: A Comprehensive Multimodal Benchmark for Theory of Mind",
    author = "Villa-Cueva, Emilio  and
      Ahmed, S M Masrur  and
      Chevi, Rendi  and
      Cruz, Jan Christian Blaise  and
      Elzeky, Kareem  and
      Cristobal, Fermin  and
      Aji, Alham Fikri  and
      Wang, Skyler  and
      Mihalcea, Rada  and
      Solorio, Thamar",
    editor = "Christodoulopoulos, Christos  and
      Chakraborty, Tanmoy  and
      Rose, Carolyn  and
      Peng, Violet",
    booktitle = "Findings of the Association for Computational Linguistics: EMNLP 2025",
    month = nov,
    year = "2025",
    address = "Suzhou, China",
    publisher = "Association for Computational Linguistics",
    url = "https://aclanthology.org/2025.findings-emnlp.1230/",
    pages = "22591--22611",
    ISBN = "979-8-89176-335-7"
}

@article{zhang2025videollama,
  title={Videollama 3: Frontier multimodal foundation models for image and video understanding},
  author={Zhang, Boqiang and Li, Kehan and Cheng, Zesen and Hu, Zhiqiang and Yuan, Yuqian and Chen, Guanzheng and Leng, Sicong and Jiang, Yuming and Zhang, Hang and Li, Xin and others},
  journal={arXiv preprint arXiv:2501.13106},
  year={2025}
}

@article{wang2025internvl3,
  title={Internvl3. 5: Advancing open-source multimodal models in versatility, reasoning, and efficiency},
  author={Wang, Weiyun and Gao, Zhangwei and Gu, Lixin and Pu, Hengjun and Cui, Long and Wei, Xingguang and Liu, Zhaoyang and Jing, Linglin and Ye, Shenglong and Shao, Jie and others},
  journal={arXiv preprint arXiv:2508.18265},
  year={2025}
}

@inproceedings{lin2023video,
    title = "Video-{LL}a{VA}: Learning United Visual Representation by Alignment Before Projection",
    author = "Lin, Bin  and
      Ye, Yang  and
      Zhu, Bin  and
      Cui, Jiaxi  and
      Ning, Munan  and
      Jin, Peng  and
      Yuan, Li",
    editor = "Al-Onaizan, Yaser  and
      Bansal, Mohit  and
      Chen, Yun-Nung",
    booktitle = "Proceedings of the 2024 Conference on Empirical Methods in Natural Language Processing",
    month = nov,
    year = "2024",
    address = "Miami, Florida, USA",
    publisher = "Association for Computational Linguistics",
    url = "https://aclanthology.org/2024.emnlp-main.342/",
    doi = "10.18653/v1/2024.emnlp-main.342",
    pages = "5971--5984"
}

@inproceedings{maaz2023video,
    title = "Video-{C}hat{GPT}: Towards Detailed Video Understanding via Large Vision and Language Models",
    author = "Maaz, Muhammad  and
      Rasheed, Hanoona  and
      Khan, Salman  and
      Khan, Fahad",
    editor = "Ku, Lun-Wei  and
      Martins, Andre  and
      Srikumar, Vivek",
    booktitle = "Proceedings of the 62nd Annual Meeting of the Association for Computational Linguistics (Volume 1: Long Papers)",
    month = aug,
    year = "2024",
    address = "Bangkok, Thailand",
    publisher = "Association for Computational Linguistics",
    url = "https://aclanthology.org/2024.acl-long.679/",
    doi = "10.18653/v1/2024.acl-long.679",
    pages = "12585--12602"
}

@article{li2025videochat,
  title={Videochat-r1: Enhancing spatio-temporal perception via reinforcement fine-tuning},
  author={Li, Xinhao and Yan, Ziang and Meng, Desen and Dong, Lu and Zeng, Xiangyu and He, Yinan and Wang, Yali and Qiao, Yu and Wang, Yi and Wang, Limin},
  journal={arXiv preprint arXiv:2504.06958},
  year={2025}
}

@inproceedings{feng2025video,
title={Video-R1: Reinforcing Video Reasoning in {MLLM}s},
author={Kaituo Feng and Kaixiong Gong and Bohao Li and Zonghao Guo and Yibing Wang and Tianshuo Peng and Junfei Wu and Xiaoying Zhang and Benyou Wang and Xiangyu Yue},
booktitle={The Thirty-ninth Annual Conference on Neural Information Processing Systems},
year={2025},
url={https://openreview.net/forum?id=a2JTVVvcEl}
}

@article{bai2025qwen2,
  title={Qwen2. 5-vl technical report},
  author={Bai, Shuai and Chen, Keqin and Liu, Xuejing and Wang, Jialin and Ge, Wenbin and Song, Sibo and Dang, Kai and Wang, Peng and Wang, Shijie and Tang, Jun and others},
  journal={arXiv preprint arXiv:2502.13923},
  year={2025}
}

@article{achiam2023gpt,
  title={Gpt-4 technical report},
  author={Achiam, Josh and Adler, Steven and Agarwal, Sandhini and Ahmad, Lama and Akkaya, Ilge and Aleman, Florencia Leoni and Almeida, Diogo and Altenschmidt, Janko and Altman, Sam and Anadkat, Shyamal and others},
  journal={arXiv preprint arXiv:2303.08774},
  year={2023}
}

@article{comanici2025gemini,
  title={Gemini 2.5: Pushing the frontier with advanced reasoning, multimodality, long context, and next generation agentic capabilities},
  author={Comanici, Gheorghe and Bieber, Eric and Schaekermann, Mike and Pasupat, Ice and Sachdeva, Noveen and Dhillon, Inderjit and Blistein, Marcel and Ram, Ori and Zhang, Dan and Rosen, Evan and others},
  journal={arXiv preprint arXiv:2507.06261},
  year={2025}
}

@article{liu2025visual,
  title={Visual-RFT: Visual Reinforcement Fine-Tuning},
  author={Liu, Ziyu and Sun, Zeyi and Zang, Yuhang and Dong, Xiaoyi and Cao, Yuhang and Duan, Haodong and Lin, Dahua and Wang, Jiaqi},
  journal={arXiv preprint arXiv:2503.01785},
  year={2025}
}

@article{luo2025robobench,
  title={Robobench: A Comprehensive Evaluation Benchmark for Multimodal Large Language Models as Embodied Brain},
  author={Luo, Yulin and Fan, Chun-Kai and Dong, Menghang and Shi, Jiayu and Zhao, Mengdi and Zhang, Bo-Wen and Chi, Cheng and Liu, Jiaming and Dai, Gaole and Zhang, Rongyu and others},
  journal={arXiv preprint arXiv:2510.17801},
  year={2025}
}

@inproceedings{zhang2024combo,
title={{COMBO}: Compositional World Models for Embodied Multi-Agent Cooperation},
author={Hongxin Zhang and Zeyuan Wang and Qiushi Lyu and Zheyuan Zhang and Sunli Chen and Tianmin Shu and Behzad Dariush and Kwonjoon Lee and Yilun Du and Chuang Gan},
booktitle={The Thirteenth International Conference on Learning Representations},
year={2025},
url={https://openreview.net/forum?id=YXRyYkb1im}
}

@inproceedings{jenamani2025feast,
  title     = {FEAST: A Flexible Mealtime-Assistance System Towards In-the-Wild Personalization},
  author    = {Rajat Kumar Jenamani and Tom Silver and Ben Dodson and Shiqin Tong and Anthony Song and 
               Yuting Yang and Ziang Liu and Benjamin Howe and Aimee Whitneck and Tapomayukh Bhattacharjee},
  booktitle = {Robotics: Science and Systems (RSS)},
  year      = {2025},
}

@misc{wang2025strangers,
      title={Communication-Efficient Desire Alignment for Embodied Agent-Human Adaptation}, 
      author={Yuanfei Wang and Xinju Huang and Fangwei Zhong and Yaodong Yang and Yizhou Wang and Yuanpei Chen and Hao Dong},
      year={2025},
      eprint={2505.22503},
      archivePrefix={arXiv},
      primaryClass={cs.RO},
      url={https://arxiv.org/abs/2505.22503}, 
}

@article{chen2023egoplan,
  title={Egoplan-bench: Benchmarking multimodal large language models for human-level planning},
  author={Chen, Yi and Ge, Yuying and Ge, Yixiao and Ding, Mingyu and Li, Bohao and Wang, Rui and Xu, Ruifeng and Shan, Ying and Liu, Xihui},
  journal={arXiv preprint arXiv:2312.06722},
  year={2023}
}

@inproceedings{sermanet2024robovqa,
  title={Robovqa: Multimodal long-horizon reasoning for robotics},
  author={Sermanet, Pierre and Ding, Tianli and Zhao, Jeffrey and Xia, Fei and Dwibedi, Debidatta and Gopalakrishnan, Keerthana and Chan, Christine and Dulac-Arnold, Gabriel and Maddineni, Sharath and Joshi, Nikhil J and others},
  booktitle={2024 IEEE International Conference on Robotics and Automation (ICRA)},
  pages={645--652},
  year={2024},
  organization={IEEE}
}

@article{li2024mmro,
title={Benchmarking Multimodal LLMs for In-Home Robotics},
author={Li, Jinming and Zhu, Yichen and Zhu, Minjie and Xu, Zhiyuan and others},
year={2024}
}

@inproceedings{cheng2024egothink,
  title={Egothink: Evaluating first-person perspective thinking capability of vision-language models},
  author={Cheng, Sijie and Guo, Zhicheng and Wu, Jingwen and Fang, Kechen and Li, Peng and Liu, Huaping and Liu, Yang},
  booktitle={Proceedings of the IEEE/CVF Conference on Computer Vision and Pattern Recognition},
  pages={14291--14302},
  year={2024}
}

@book{baron1997mindblindness,
  title={Mindblindness: An essay on autism and theory of mind},
  author={Baron-Cohen, Simon},
  year={1997},
  publisher={MIT press}
}

@InProceedings{Gao_2025_ACMMM,
        author    = {Junyu Gao, Xuan Yao, Yong Rui and Changsheng Xu},
        title     = {Building Embodied EvoAgent: A Brain-inspired Paradigm for Bridging Multimodal Large Models and World Models},
        booktitle = {Proceedings of the 33rd ACM International Conference on Multimedia (ACM MM)},
        year      = {2025},
    }

@inproceedings{Zhang2019BERTScore,
title={BERTScore: Evaluating Text Generation with BERT},
author={Tianyi Zhang* and Varsha Kishore* and Felix Wu* and Kilian Q. Weinberger and Yoav Artzi},
booktitle={International Conference on Learning Representations},
year={2020},
url={https://openreview.net/forum?id=SkeHuCVFDr}
}

@inproceedings{reimers-2020-multilingual-sentence-bert,
    title = "Making Monolingual Sentence Embeddings Multilingual using Knowledge Distillation",
    author = "Reimers, Nils and Gurevych, Iryna",
    booktitle = "Proceedings of the 2020 Conference on Empirical Methods in Natural Language Processing",
    month = "11",
    year = "2020",
    publisher = "Association for Computational Linguistics",
    url = "https://arxiv.org/abs/2004.09813",
}

@article{li2024llava,
  title={LLaVA-OneVision: Easy Visual Task Transfer},
  author={Li, Bo and Zhang, Yuanhan and Guo, Dong and Zhang, Renrui and Li, Feng and Zhang, Hao and Zhang, Kaichen and Zhang, Peiyuan and Li, Yanwei and Liu, Ziwei and Li, Chunyuan},
  journal={Transactions on Machine Learning Research},
  year={2024}
}

@article{fung2025embodied,
  title={Embodied ai agents: Modeling the world},
  author={Fung, Pascale and Bachrach, Yoram and Celikyilmaz, Asli and Chaudhuri, Kamalika and Chen, Delong and Chung, Willy and Dupoux, Emmanuel and Gong, Hongyu and J{\'e}gou, Herv{\'e} and Lazaric, Alessandro and others},
  journal={arXiv preprint arXiv:2506.22355},
  year={2025}
}

@inproceedings{yangembodiedbench,
  title={EmbodiedBench: Comprehensive Benchmarking Multi-modal Large Language Models for Vision-Driven Embodied Agents},
  author={Yang, Rui and Chen, Hanyang and Zhang, Junyu and Zhao, Mark and Qian, Cheng and Wang, Kangrui and Wang, Qineng and Koripella, Teja Venkat and Movahedi, Marziyeh and Li, Manling and others},
  booktitle={Forty-second International Conference on Machine Learning}
}

@inproceedings{xu2025llava,
  title={Llava-cot: Let vision language models reason step-by-step},
  author={Xu, Guowei and Jin, Peng and Wu, Ziang and Li, Hao and Song, Yibing and Sun, Lichao and Yuan, Li},
  booktitle={Proceedings of the IEEE/CVF International Conference on Computer Vision},
  pages={2087--2098},
  year={2025}
}

@article{yang2025qwen3,
  title={Qwen3 technical report},
  author={Yang, An and Li, Anfeng and Yang, Baosong and Zhang, Beichen and Hui, Binyuan and Zheng, Bo and Yu, Bowen and Gao, Chang and Huang, Chengen and Lv, Chenxu and others},
  journal={arXiv preprint arXiv:2505.09388},
  year={2025}
}

@article{leslie2004core,
  title={Core mechanisms in ‘theory of mind’},
  author={Leslie, Alan M and Friedman, Ori and German, Tim P},
  journal={Trends in cognitive sciences},
  volume={8},
  number={12},
  pages={528--533},
  year={2004},
  publisher={Elsevier}
}

@article{onishi200515,
  title={Do 15-month-old infants understand false beliefs?},
  author={Onishi, Kristine H and Baillargeon, Ren{\'e}e},
  journal={science},
  volume={308},
  number={5719},
  pages={255--258},
  year={2005},
  publisher={American Association for the Advancement of Science}
}

@article{frith2005theory,
  title={Theory of mind},
  author={Frith, Chris and Frith, Uta},
  journal={Current biology},
  volume={15},
  number={17},
  pages={R644--R645},
  year={2005},
  publisher={Elsevier}
}

@inproceedings{zhang2025autotom,
  title={Autotom: Automated bayesian inverse planning and model discovery for open-ended theory of mind},
  author={Zhang, Zhining and Jin, Chuanyang and Jia, Mung Yao and Shu, Tianmin},
  booktitle={ICLR 2025 Workshop on Foundation Models in the Wild},
  year={2025}
}

@inproceedings{wilf2024think,
  title={Think twice: Perspective-taking improves large language models’ theory-of-mind capabilities},
  author={Wilf, Alex and Lee, Sihyun and Liang, Paul Pu and Morency, Louis-Philippe},
  booktitle={Proceedings of the 62nd Annual Meeting of the Association for Computational Linguistics (Volume 1: Long Papers)},
  pages={8292--8308},
  year={2024}
}

@inproceedings{jung2024perceptions,
    title = "Perceptions to Beliefs: Exploring Precursory Inferences for Theory of Mind in Large Language Models",
    author = "Jung, Chani  and
      Kim, Dongkwan  and
      Jin, Jiho  and
      Kim, Jiseon  and
      Seonwoo, Yeon  and
      Choi, Yejin  and
      Oh, Alice  and
      Kim, Hyunwoo",
    editor = "Al-Onaizan, Yaser  and
      Bansal, Mohit  and
      Chen, Yun-Nung",
    booktitle = "Proceedings of the 2024 Conference on Empirical Methods in Natural Language Processing",
    month = nov,
    year = "2024",
    address = "Miami, Florida, USA",
    publisher = "Association for Computational Linguistics",
    url = "https://aclanthology.org/2024.emnlp-main.1105/",
    doi = "10.18653/v1/2024.emnlp-main.1105",
    pages = "19794--19809"
}

@inproceedings{zhang2025vlabench,
  title={Vlabench: A large-scale benchmark for language-conditioned robotics manipulation with long-horizon reasoning tasks},
  author={Zhang, Shiduo and Xu, Zhe and Liu, Peiju and Yu, Xiaopeng and Li, Yuan and Gao, Qinghui and Fei, Zhaoye and Yin, Zhangyue and Wu, Zuxuan and Jiang, Yu-Gang and others},
  booktitle={Proceedings of the IEEE/CVF International Conference on Computer Vision},
  pages={11142--11152},
  year={2025}
}

@inproceedings{palme,
author = {Driess, Danny and Xia, Fei and Sajjadi, Mehdi S. M. and Lynch, Corey and Chowdhery, Aakanksha and Ichter, Brian and Wahid, Ayzaan and Tompson, Jonathan and Vuong, Quan and Yu, Tianhe and Huang, Wenlong and Chebotar, Yevgen and Sermanet, Pierre and Duckworth, Daniel and Levine, Sergey and Vanhoucke, Vincent and Hausman, Karol and Toussaint, Marc and Greff, Klaus and Zeng, Andy and Mordatch, Igor and Florence, Pete},
title = {PaLM-E: an embodied multimodal language model},
year = {2023},
publisher = {JMLR.org},
booktitle = {Proceedings of the 40th International Conference on Machine Learning},
articleno = {340},
numpages = {20},
location = {Honolulu, Hawaii, USA},
series = {ICML'23}
}

@article{shao2024deepseekmath,
  title={Deepseekmath: Pushing the limits of mathematical reasoning in open language models},
  author={Shao, Zhihong and Wang, Peiyi and Zhu, Qihao and Xu, Runxin and Song, Junxiao and Bi, Xiao and Zhang, Haowei and Zhang, Mingchuan and Li, YK and Wu, Yang and others},
  journal={arXiv preprint arXiv:2402.03300},
  year={2024}
}
